\pdfoutput=1
\documentclass[letterpaper,journal]{IEEEtran}
\usepackage{amsmath,amsfonts}
\usepackage{array}
\usepackage[caption=false,font=normalsize,labelfont=sf,textfont=sf]{subfig}
\usepackage{textcomp}
\usepackage{stfloats}
\usepackage{url}
\usepackage{graphicx}
\usepackage{xcolor}
\usepackage{cite}
\usepackage[hidelinks]{hyperref}

\DeclareRobustCommand{\authorlink}[3]{\hyperlink{#1}{#2\textsuperscript{#3}}}
\DeclareRobustCommand{\emailanchor}[2]{\raisebox{\baselineskip}[0pt][0pt]{\hypertarget{#1}{}}\textsuperscript{#2}}



\begin{document}

\title{GeoPillars: A Large-Kernel Pillar Detector for Single-Sweep LiDAR Under Point Sparsity, and a Decomposition of Its Gain}

\author{\authorlink{email:dumbre}{Chaitanya~Dumbre}{*},~
  \authorlink{email:sarkar}{Shubham~Sarkar}{\textdagger},~
  \authorlink{email:mohapatra}{Akash~Mohapatra}{\textdaggerdbl},~
  \authorlink{email:patil}{Ghanasham~Patil}{\S},~
  and~\authorlink{email:girase}{Sheetal~Girase}{\P}%
  \thanks{Preprint, November 2026.}%
  \thanks{C.~Dumbre, S.~Sarkar, A.~Mohapatra, G.~Patil, and S.~Girase are with the
    School of Computer Science and Engineering, MIT World Peace University,
    Pune, India
    (e-mail: \emailanchor{email:dumbre}{*}chaidumb04@gmail.com; \emailanchor{email:sarkar}{\textdagger}shubham04sarkar@gmail.com;
    \emailanchor{email:mohapatra}{\textdaggerdbl}akashmahapatra762@gmail.com; \emailanchor{email:patil}{\S}ghpatil27@gmail.com;
    \emailanchor{email:girase}{\P}sheetal.girase@mitwpu.edu.in).}}

\markboth{Preprint}%
{Dumbre \MakeLowercase{\textit{et al.}}: GeoPillars, a Large-Kernel Pillar Detector for Single-Sweep LiDAR}

\maketitle

\begin{abstract}
  GeoPillars is a pillar-based 3D detector for single-sweep LiDAR under a hard budget of eight points per 0.4~m pillar, a budget that discards 46.6\% of in-range points. It adds geometric pillar features, grouped vector attention within each pillar, a learned absolute positional prior on the bird's-eye-view (BEV) grid, a large-kernel backbone, auxiliary surface-normal supervision and a full-resolution detection head. Trained on 123k synthetic CARLA scans and evaluated on cars, it reaches 81.33 BEV AP@0.7, 13.76 points above CenterPoint at the same input budget and 1.91 points above CenterPoint at its native, denser budget. These are in-distribution figures, because the validation split is per-frame rather than scene-disjoint. A rung-by-rung decomposition shows that the new components explain little of the gain. Raising the detection-head resolution alone recovers 10.19 of the 13.76 points, the joint change of encoder, backbone, neck and positional prior adds 3.37, and the normal-supervision branch adds 0.20, which is within run-to-run variation. With heading reversals folded out, orientation error is similar for all seven models, although GeoPillars reverses headings about half as often as a resolution-matched baseline. It is also the slowest of the seven models. Evaluated zero-shot on 6,743 real nuScenes keyframes, no model transfers, and each retains 3.7\% to 12.2\% of its CARLA accuracy. A collection defect left LiDAR intensity constant, so every model learned from geometry alone. Code and checkpoints are publicly available.
\end{abstract}


\begin{figure}[!t]
  \centering
  \includegraphics[width=\columnwidth]{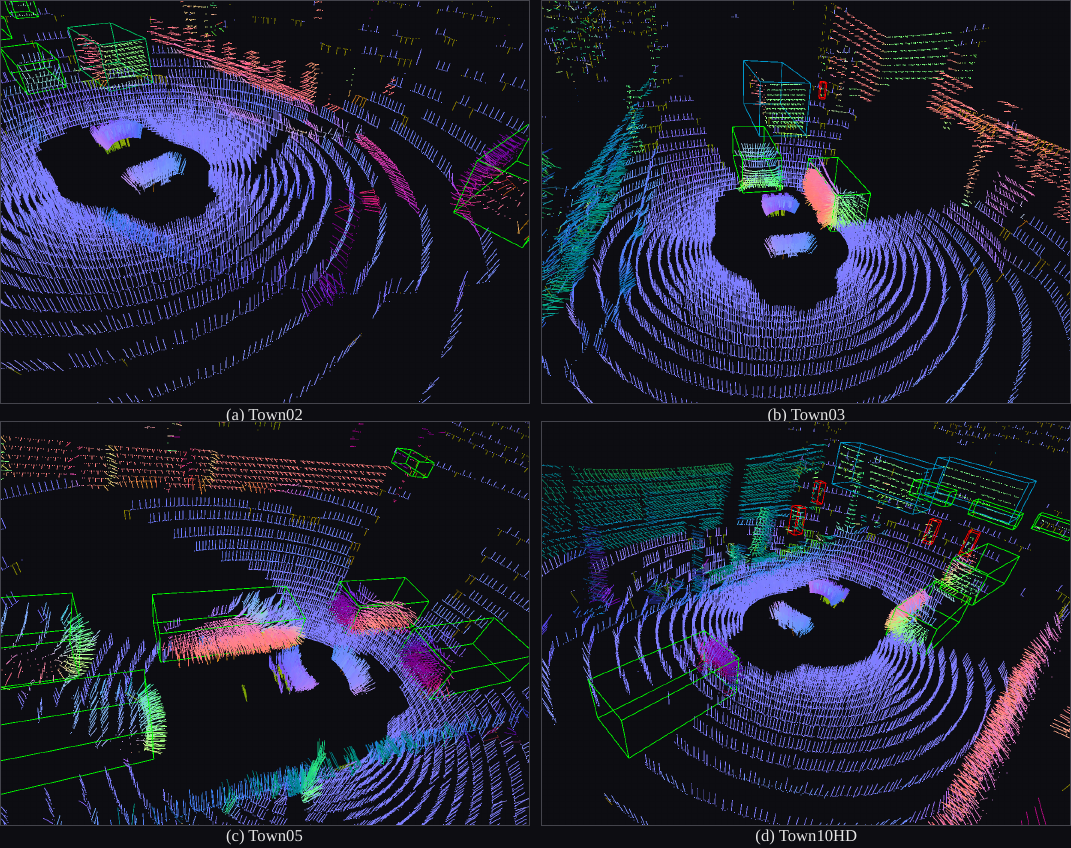}
  \caption{Surface normals estimated offline and used as the auxiliary target, for one scan from each CARLA town, (a) Town02, (b) Town03, (c) Town05 and (d) Town10HD. Each point is drawn with its unit normal coloured by direction, and ground-truth boxes are green for cars, cyan for buses and red for pedestrians. The normals separate vehicle surfaces from the ground, but 91.4\% of the auxiliary loss weight falls on patches with no object (Section~\ref{subsec:audit}).}
  \label{fig:normals_teaser}
\end{figure}

\begin{figure*}[!t]
  \centering
  \includegraphics[width=\textwidth]{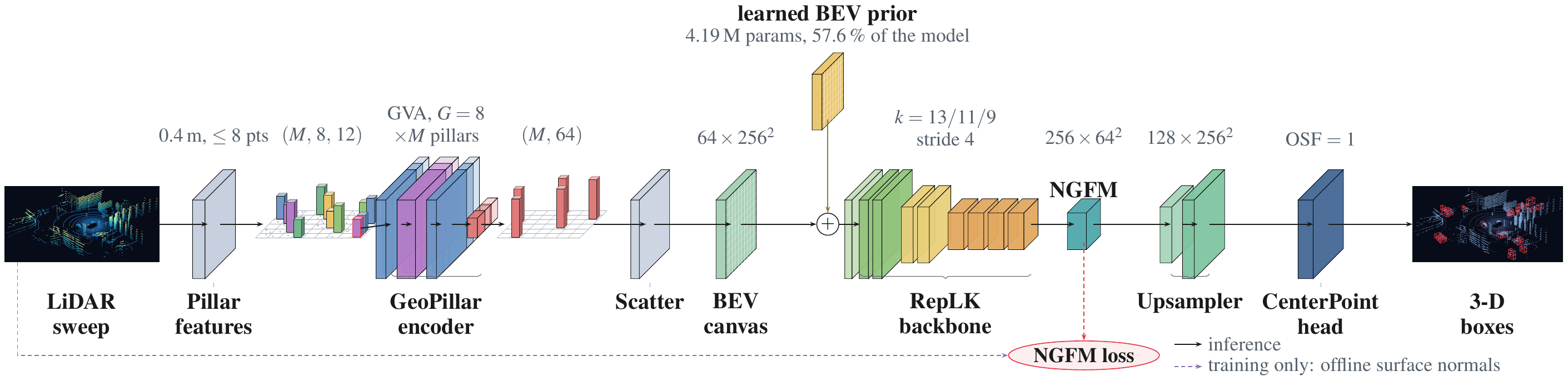}
  \caption{GeoPillars. Each 0.4~m pillar keeps at most eight points, and the builder appends eight geometric features (cluster offset, pillar-centre offset, range, azimuth and density) to give 12 per point. The encoder mixes the points of a pillar with grouped vector attention ($G=8$), max-pools them to a 64-dimensional vector, scatters the result to a $256\times256$ canvas and adds the learned positional prior $E_{\text{pos}}\in\mathbb{R}^{1\times64\times256\times256}$, which holds 4.19~M parameters, 57.6\% of the model. A three-stage large-kernel backbone ($k=13/11/9$ with parallel $5\times5$ branches, total stride 4) reduces the canvas to $256\times64^2$, where NGFM predicts a unit normal per cell and gates the features by $x\odot(1+\sigma(\mathrm{BN}(W\hat{n})))$. Two transposed-convolution stages restore $128\times256^2$ for a stock CenterPoint head at \texttt{out\_size\_factor}~$=1$. The dashed path is used only in training. Offline normals from the same sweep are pooled to $64^2$ and supervise $\hat{n}$ through a masked $1-\cos(\hat{n},n)$ loss on cells whose occupancy exceeds 10\% of a $4\times4$ patch. The end panels show a CARLA validation scan and its epoch-8 detections at score $\ge0.25$.}
  \label{fig:architecture}
\end{figure*}

\section{Introduction}

\IEEEPARstart{P}{illar}-based detectors~\cite{lang2019pointpillars} encode a LiDAR sweep as vertical columns on a bird's-eye-view (BEV) grid, and they have little to work with when the sweep comes from a single low-channel sensor. The sparse regime studied here is self-imposed. Each 0.4~m pillar of our 32-channel sweeps keeps at most eight points, so hard voxelisation~\cite{pmlr-v100-zhou20a} discards 46.6\% of the in-range points before the encoder sees them. The cap does not come from any sensor. It is a controlled degradation, applied to every run except one baseline (Section~\ref{subsec:sparsity}). Standard pillar encoders describe each cell by per-point offsets from the cluster mean~\cite{2018voxelnet, lang2019pointpillars} and from the pillar centre~\cite{lang2019pointpillars}. With eight points, we expected these offsets to say little about local surface orientation, leaving the detector to infer the extent and heading of a vehicle from a sparse, partial shell.

We designed GeoPillars to compensate in four ways. First, the pillar builder computes range, azimuth and occupancy density in closed form, so the encoder does not have to learn them from eight points. Second, the encoder relates the points of each pillar through a modified grouped vector attention (GVA) block~\cite{wu2022pointtrasnformerV2}. The simplified PointNet of PointPillars transforms every point independently before max-pooling~\cite{qi2017pointnet, lang2019pointpillars}, and we expected attention over the relative arrangement of the points to learn local geometric structure that independent transforms cannot. Third, large-kernel convolutions~\cite{chen2023largekernel3d, 2022largekernel} on the BEV grid let a single layer span a whole vehicle footprint and gather evidence across neighbouring pillars. Fourth, an auxiliary branch predicts surface normals estimated offline from the full-density cloud~\cite{Hoppe1992Surface} (Fig.~\ref{fig:normals_teaser}), in the spirit of learning using privileged information (LUPI)~\cite{VAPNIK2009544} and of the training-only auxiliary network of SA-SSD~\cite{He2020_SASSD}. This supervision was meant to shape features that encode surface orientation. Most of these expectations did not survive the evaluation.

Fig.~\ref{fig:architecture} shows the resulting system. The encoder output is scattered onto a $256 \times 256$ BEV grid~\cite{lang2019pointpillars} and summed with a learned bias for every grid cell. A three-stage large-kernel backbone, adapted from the CenterPoint backbone of LargeKernel3D~\cite{chen2023largekernel3d} with blocks in the style of RepLKNet~\cite{2022largekernel}, reduces the grid to $64 \times 64$, where the normal-guided branch acts, and an upsampler restores full resolution for a CenterPoint head~\cite{yin2021center}. The upsampler was added together with the normal branch but does not depend on it, and the head resolution it provides turned out to matter more than any other change.

We trained GeoPillars, two ablations of it and four CenterPoint~\cite{yin2021center} baselines on 123,364 synthetic CARLA~\cite{dosovitskiy2017carla} scans collected to the nuScenes~\cite{caesar2020nuscenes} sensor specification, and scored all seven on cars. Table~\ref{tab:model_zoo} lists the runs. Run A is GeoPillars and B is A without the intensity feature. C is CenterPoint at its native input budget of 0.2~m pillars with up to 20 points~\cite{yin2021center, mmdet3d2020}, and D is CenterPoint at the GeoPillars budget. E is D without the intensity feature, F is D with a full-resolution head, and G is A without the normal-guided branch. All runs are evaluated on a 30,842-scan CARLA validation split and, zero-shot, on 6,743 real nuScenes keyframes. The CARLA split is per-frame rather than scene-disjoint, a form of train--test leakage that makes held-out figures optimistic~\cite{KAPOOR2023100804}, so every CARLA number in this paper measures in-distribution interpolation.

GeoPillars reaches 81.33 AP@0.7, 13.76 points above D and 1.91 points above C. Section~\ref{subsec:decomposition} takes the 13.76-point gap apart along the ladder D$\to$F$\to$G$\to$A, one run at a time. Raising the head resolution of the baseline (D$\to$F) recovers 10.19 points on its own. Replacing the encoder, backbone and neck (F$\to$G) adds 3.37 points, but this rung also narrows the head input, adds the positional bias and changes the batch-normalisation statistics, and the intermediate runs that would separate these changes were not trained. The normal branch (G$\to$A) adds 0.20 points, which is inside the measured run-to-run variation. The architecture does reverse vehicle headings about half as often as F, yet it is the slowest model in the study, and none of the seven models transfers to nuScenes (Section~\ref{subsec:zeroshot}). Section~\ref{sec:discussion} draws five lessons from these outcomes that we expect to apply beyond this network.

This paper makes four contributions.
\begin{itemize}
  \item A rung-by-rung decomposition of a 13.76-point detection gain, in which the one rung that changes a single model setting accounts for 74\% of it (Section~\ref{subsec:decomposition}).
  \item Measurements of what the attention block, the positional prior and the normal-guided gate do in the trained checkpoints, set against their design intent.
  \item Three negative results, on auxiliary normal supervision, synthetic-to-real transfer and the inference cost of a design motivated by sparsity.
  \item An implementation audit (Section~\ref{subsec:audit}), released together with the collection pipeline, configurations, checkpoints and evaluation records.
\end{itemize}

All results are single-class and single-run, and every model learned from geometry alone, because a collection defect left LiDAR intensity constant (Section~\ref{subsec:audit}). The absolute numbers should therefore not be read as a benchmark. The comparisons between models remain internally valid, since all seven share the same data, schedule, augmentation and evaluator.

\begin{table*}[!t]
  \caption{The seven runs. The encoder is the \texttt{PillarFeatureNet} of PointPillars~\cite{lang2019pointpillars} or the GeoPillar encoder of Section~\ref{subsec:prior}. SECOND and SECONDFPN are the backbone and neck of~\cite{yan2018second}, and RepLK is the backbone of Section~\ref{subsec:backbone}. OS is the output stride of the detection head relative to the pillar grid. Figures label it OSF, after the \texttt{out\_size\_factor} setting, and abbreviate CenterPoint as CP. In Dim is the configured encoder input dimension. It includes the eight appended geometric features for the GeoPillar encoder, whereas \texttt{PillarFeatureNet} appends six offset features internally. Parameter counts in parentheses exclude the positional bias. Train Mem is the median per-GPU peak allocation in the training log and is not comparable across GPU types.}
  \label{tab:model_zoo}
  \centering
  \setlength{\tabcolsep}{4pt}
  \begin{tabular}{c l l l c c c c c c c c}
    \hline
    Run & Encoder          & Backbone & Neck             & OS & Voxel / Pts / Cap & In Dim & Params          & Prior & $\beta$ & GPU            & Train Mem \\
    \hline
    A   & GeoPillar        & RepLK    & NGFM + upsampler & 1  & 0.4 / 8 / 16k     & 12     & 7.28~M (3.08~M) & yes   & 0.44    & H100           & 53.5~GiB  \\
    B   & GeoPillar        & RepLK    & NGFM + upsampler & 1  & 0.4 / 8 / 16k     & 11     & 7.28~M (3.08~M) & yes   & 0.44    & H100           & 53.5~GiB  \\
    C   & PillarFeatureNet & SECOND   & SECONDFPN        & 4  & 0.2 / 20 / 30k    & 4      & 4.84~M          & no    & 0.11    & A100           & 30.5~GiB  \\
    D   & PillarFeatureNet & SECOND   & SECONDFPN        & 4  & 0.4 / 8 / 16k     & 4      & 4.84~M          & no    & 0.44    & H100           & 7.7~GiB   \\
    E   & PillarFeatureNet & SECOND   & SECONDFPN        & 4  & 0.4 / 8 / 16k     & 3      & 4.84~M          & no    & 0.44    & A10            & 7.6~GiB   \\
    F   & PillarFeatureNet & SECOND   & SECONDFPN        & 1  & 0.4 / 8 / 16k     & 4      & 7.05~M          & no    & 0.44    & A100           & 35.8~GiB  \\
    G   & GeoPillar        & RepLK    & upsampler        & 1  & 0.4 / 8 / 16k     & 12     & 6.98~M (2.78~M) & yes   & 0.44    & 2$\times$ 5090 & 26.2~GiB  \\
    \hline
  \end{tabular}
\end{table*}

\section{Related Work}
PointPillars~\cite{lang2019pointpillars} encodes vertical columns of points with a simplified PointNet~\cite{qi2017pointnet} into a BEV pseudo-image that 2D convolutions process efficiently, avoiding the 3D convolutions of voxel detectors such as VoxelNet~\cite{2018voxelnet} and SECOND~\cite{yan2018second}. CenterPoint~\cite{yin2021center} replaced anchor boxes with centre-based keypoint estimation~\cite{zhou2019objects}, and its pillar variant runs on a 0.2~m grid on nuScenes~\cite{caesar2020nuscenes}. Pillar detectors usually downsample the canvas through strided stages~\cite{lang2019pointpillars, yin2021center}, and output resolution is known to matter. SST~\cite{2022SST} showed in a pilot study that network stride affects LiDAR detectors, and PillarNeXt~\cite{2023PillarNeXt} found that an upsampling layer in the detection head gives a clear gain, especially for small objects. We apply this control to a component comparison by changing only the head stride on one rung.

Within a pillar, standard encoders summarise geometry with per-point coordinate offsets~\cite{2018voxelnet, lang2019pointpillars}. Vector attention instead computes a weight per feature dimension from a relation between query and key features, such as their difference~\cite{zhao2020san}, and Point Transformer~\cite{zhao2021point} applied it to point clouds. GeoPillars uses a modified form of the grouped vector attention (GVA) of Point Transformer V2~\cite{wu2022pointtrasnformerV2}, in which groups of feature dimensions share attention weights, without its weight-encoding layer or positional encoding. For context across cells, LargeKernel3D~\cite{chen2023largekernel3d} showed that 3D sparse convolutional backbones have limited effective receptive fields because sparse features are spatially disconnected, and it addressed this with a spatial-wise partition convolution, scaling kernels up to $17\times17\times17$ on Waymo~\cite{sun2020waymo}. In 2D, RepLKNet~\cite{2022largekernel} argued that a few large re-parameterised depthwise kernels outperform a stack of small ones. Our backbone carries the CenterPoint layout of LargeKernel3D over to a dense BEV canvas and builds its blocks in the style of RepLKNet.

LUPI~\cite{VAPNIK2009544} formalises training with features that are available during optimisation but not at test time. In LiDAR detection, SA-SSD~\cite{He2020_SASSD} trains a detachable auxiliary network with point-level supervision to make backbone features structure-aware. Explicit geometric features have a longer history in depth-based detection. HHA~\cite{gupta2014learning} encoded depth as horizontal disparity, height above ground and angle with gravity, and learned better CNN features than raw depth did. Deep Sliding Shapes~\cite{song2016deep}, however, found that adding HHA to a network that already operates on a directional TSDF volume gave no benefit, because the 3D representation already captured that information. Our auxiliary normals raise the same concern, since they are computed from the sweep the network receives.

Sliding Shapes~\cite{song2014sliding} trained 3D detectors on depth maps rendered from CAD models, and simulators such as GTA-V and CARLA~\cite{dosovitskiy2017carla} have since been used to generate labelled LiDAR scans without manual annotation~\cite{wu2019squeezesegv2, 2019presil, manuallabelfree2020}. Models trained this way show large domain gaps on real data~\cite{wu2019squeezesegv2, manuallabelfree2020}, and point density~\cite{wei2022lidar}, ray drop~\cite{wu2019squeezesegv2, manivasagam2020lidarsim, guillard2022learning} and intensity~\cite{wu2019squeezesegv2, guillard2022learning} have all been reported to contribute. In our zero-shot evaluation the intensity term is absent by construction, because the training data carry a constant intensity and the nuScenes intensity is overwritten with the same constant. Geometry, density and label definition remain as candidate causes of the transfer failure.

\section{System Design and Rationale}
\label{sec:design}
This section describes GeoPillars as built and the reasoning behind each component.

\subsection{Pillar feature construction}
\label{subsec:pillar}
\begin{figure}[!t]
  \centering
  \includegraphics[width=\columnwidth]{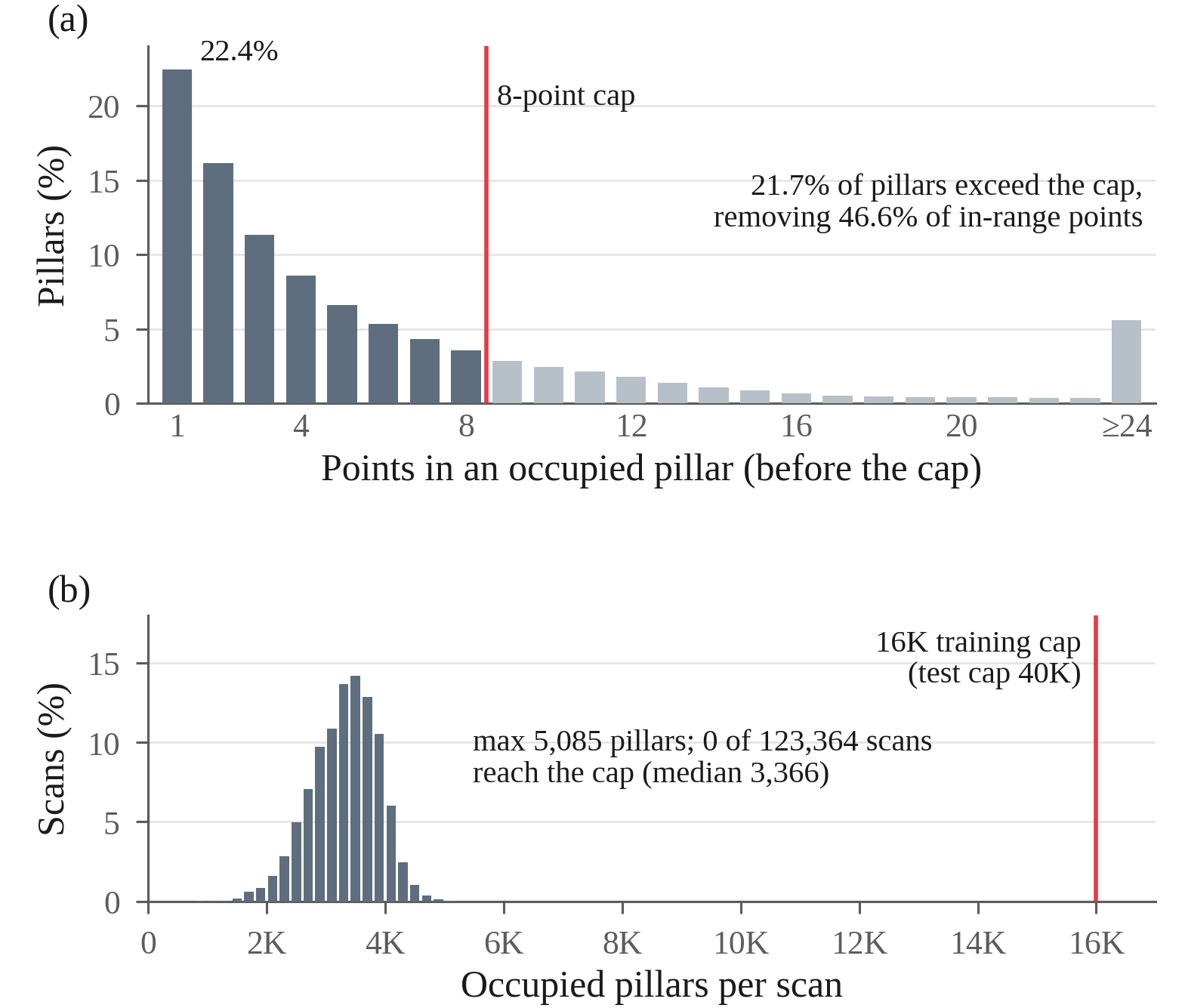}
  \caption{The sparsity regime over the 123,364 training scans. (a) Points per occupied pillar before the cap. 21.7\% of occupied pillars exceed eight points, and discarding the surplus removes 46.6\% of in-range points. (b) Occupied pillars per scan. The densest scan occupies 5,085 pillars, so the 16,000-pillar cap never activates and the point cap is the only constraint that binds.}
  \label{fig:pillar_stats}
\end{figure}

Points are assigned to a $256 \times 256$ grid of $0.4$~m pillars spanning $\pm 51.2$~m. The assignment uses the hard voxelisation~\cite{pmlr-v100-zhou20a} of mmcv~\cite{mmcv}, which caps each pillar at a fixed capacity and fills it in arrival order. Each pillar therefore keeps its first eight points rather than a random sample~\cite{2018voxelnet, lang2019pointpillars}, and the cap discards 46.6\% of in-range points (Fig.~\ref{fig:pillar_stats}(a)). Because the training pipeline shuffles points first, a training pillar in effect keeps a random eight while inference keeps the first eight in scan order (Section~\ref{subsec:audit}). For each of the $n \le 8$ points $p = (x, y, z)$ retained in a pillar, with mean $\bar{p}$ over those $n$ points and pillar centre $(x_c, y_c)$, the encoder receives
\begin{equation}
  \label{eq:pillar_features}
  f = \begin{bmatrix}
    p                          \\
    i                          \\
    p - \bar{p}                \\
    x - x_c                    \\
    y - y_c                    \\
    \Vert p \Vert_2            \\
    \operatorname{atan2}(y, x) \\
    n / P_{\text{max}}
  \end{bmatrix} \in \mathbb{R}^{12}
\end{equation}
where $i$ is the intensity and $P_{\text{max}} = 8$ is the point cap. Padded slots are zeroed so that the tensor keeps a fixed size. Table~\ref{tab:features} names the terms of \eqref{eq:pillar_features} in order, with the source of each and the role it was expected to play.

\begin{table}[!t]
  \caption{The Terms of \eqref{eq:pillar_features}. Intensity is the only ablated term (runs B and E), so the last column states design intent rather than a measured effect.}
  \label{tab:features}
  \centering
  \setlength{\tabcolsep}{3pt}
  \begin{tabular}{l c c p{4.75cm}}
    \hline
    Term           & Dim & Source                      & Expected role                                           \\
    \hline
    Coordinates    & 3   & \cite{lang2019pointpillars} & Absolute position of the point                          \\
    Intensity      & 1   & \cite{lang2019pointpillars} & Surface reflectance, constant at 1.0 in this dataset    \\
    Cluster offset & 3   & \cite{lang2019pointpillars} & Spread of the points about their pillar mean            \\
    Centre offset  & 2   & \cite{lang2019pointpillars} & Position within the 0.4~m cell                          \\
    Range          & 1   & Ours                        & Distance, which sets the point spacing on a surface     \\
    Azimuth        & 1   & Ours                        & Direction of the ray that sampled the surface           \\
    Density        & 1   & Ours                        & Separates a full pillar from one holding a single point \\
    \hline
  \end{tabular}
\end{table}

\subsection{Grouped Vector Attention Encoder and the Positional Prior}
\label{subsec:prior}

\begin{figure*}[!t]
  \centering
  \includegraphics[width=\textwidth]{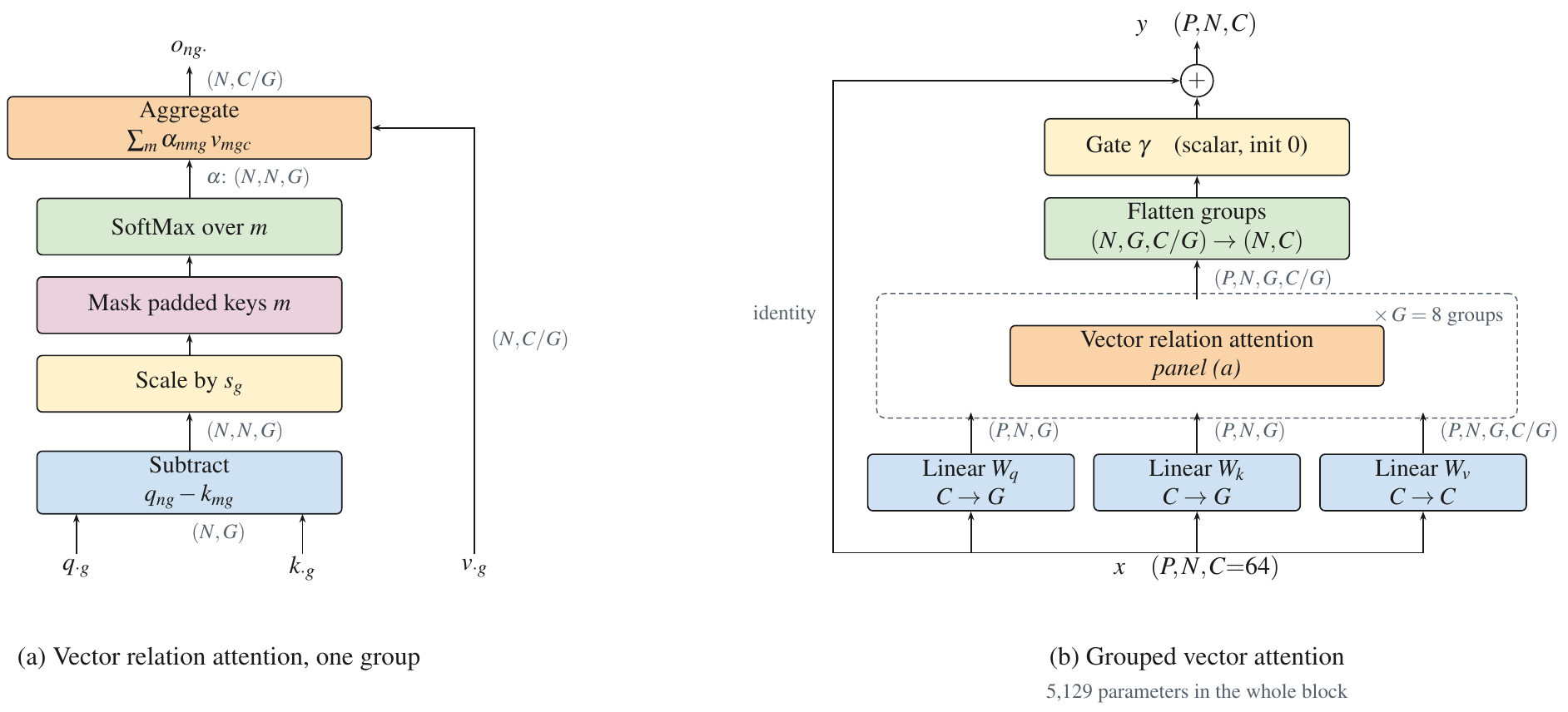}
  \caption{Grouped vector attention. (a)~One group. The logit is the \emph{difference} of two low-rank projections, $q_{ng}-k_{mg}$, so the block scores relative arrangement rather than agreement. The per-group scale $s_g$ is the only temperature, and both the mask and the softmax act on the key axis $m$. (b)~$G=8$ groups over $C=64$ dimensions, with each attention map shared by $C/G=8$ dimensions, so the query and key projections hold 512 parameters each and the block 5{,}129. The residual gate $\gamma$ is a scalar initialised to zero, and after eight epochs $|\gamma| \le 0.038$ in every run that carries the block. $P$ is the number of occupied pillars and $N = 8$ the number of point slots per pillar.}
  \label{fig:gva}
\end{figure*}

Each point's 12 features are lifted to 64 dimensions by a linear layer with batch normalisation~\cite{pmlr2015batchnorm} and ReLU. A modified grouped vector attention block~\cite{wu2022pointtrasnformerV2} then mixes the points of the pillar (Fig.~\ref{fig:gva}), and a second linear layer with batch normalisation and ReLU precedes a max-pool over the valid points~\cite{qi2017pointnet}. With $G = 8$ groups over $C = 64$ dimensions, for query point $n$, key point $m$ and group $g$,
\begin{align*}
  q_{ng}        & = \sum_c W_q[g,c] x_{nc}                                \\
  k_{mg}        & = \sum_c W_k[g,c] x_{mc}                                \\
  e_{nmg}       & = (q_{ng} - k_{mg}) \cdot s_g                           \\
  \hat{e}_{nmg} & = -10^4 \text{ if } m \text{ is padding, else } e_{nmg} \\
  \alpha_{nmg}  & = \text{softmax}_m(\hat{e}_{nmg})                       \\
  v             & = \text{reshape}(W_v x) \in \mathbb{R}^{G \times (C/G)} \\
  o_{ngc}       & = \sum_m \alpha_{nmg} v_{mgc}                           \\
  y_n           & = x_n + \gamma \cdot \text{flatten}(o_n)
\end{align*}

The logit uses the subtraction relation of vector attention~\cite{zhao2020san} rather than the dot product of a transformer~\cite{vaswani2017attention}, so it scores the relative arrangement of two points. The block is meant to let the relations between the points of a pillar carry local surface geometry. The block departs from the GVA of Point Transformer V2 in four ways. Queries and keys are projected straight to the $G$ groups, so their scaled difference is the logit and no weight-encoding layer follows it. There is no positional-encoding multiplier or bias, so position enters only through the coordinates and offsets of \eqref{eq:pillar_features}. Attention runs over the at most eight points of one pillar, with padded keys masked, rather than over a $k$-nearest-neighbour set. Finally, the output passes through a scalar residual gate. The gate $\gamma$ starts at zero, as in ReZero~\cite{bachlechner2021rezero}, so the block begins as the identity, and it stays close to the identity in training. After eight epochs $|\gamma| \le 0.038$ in all three runs that carry the block, and in run A the block contributes 1.8\% of the feature magnitude, spread almost evenly across points. The trained encoder therefore behaves as two linear layers with batch normalisation and ReLU followed by a masked max-pool, which yields one 64-dimensional vector per pillar. We did not train an encoder without the block, so this describes the trained block and not whether training needs it.

The encoder scatters these vectors onto the BEV canvas~\cite{lang2019pointpillars} and adds a learned tensor $E_{\text{pos}} \in \mathbb{R}^{1 \times 64 \times 256 \times 256}$, one bias per cell and channel, initialised from a zero-mean Gaussian with standard deviation 0.01. It was added as a coarse signal of absolute position, on the reasoning that a detector working in the ego frame benefits from knowing roughly where a feature sits. Four of its properties matter for the results.

\begin{enumerate}
  \item \emph{Size.} It holds 4,194,304 parameters, 57.6\% of the model's 7,276,116, and without it GeoPillars has fewer parameters (3.08~M) than the 4.84~M CenterPoint baseline. It acts mainly as a fill-in for empty space. It supplies all of the content of the 94.9\% of cells that are exactly zero before it is added. In a forward pass over eight validation frames, however, it makes up only 0.35\% of the map's $\ell^2$ energy overall and 2.3\% of its mean absolute value in occupied cells. Section~\ref{subsec:prior_ablation} measures its effect on accuracy.

  \item \emph{Absolute indexing.} It is indexed by absolute ego-frame position, so it could encode where objects tend to appear in the training towns rather than any transferable geometry. That would explain the poor transfer to nuScenes, but the evidence points the other way (Section~\ref{subsec:pm_prior}).

  \item \emph{No baseline counterpart.} The \texttt{PillarFeatureNet} of the baselines~\cite{mmdet3d2020} holds 768 parameters (704 in run E) and \texttt{PointPillarsScatter} holds none, so every comparison involving the encoder swap, including the F$\to$G rung, also includes this bias.

  \item \emph{No empty cells.} Because the bias is added to every cell, no cell of the encoder output is exactly zero. This disables a density weight in the auxiliary loss (Section~\ref{subsec:ngfm}).
\end{enumerate}

\subsection{Large-kernel bird's-eye-view backbone}
\label{subsec:backbone}
The backbone adapts the CenterPoint backbone of LargeKernel3D~\cite{chen2023largekernel3d} to the dense 2D BEV canvas. It keeps that network's layout, a stem followed by stages of residual blocks with a strided convolution between stages, and replaces its sparse 3D blocks with depthwise large-kernel blocks in the style of RepLKNet~\cite{2022largekernel}. A $3 \times 3$ stem keeps 64 channels at full resolution. Three stages follow, with two blocks of 64 channels and a $13 \times 13$ depthwise kernel at $256 \times 256$, two blocks of 128 channels and an $11 \times 11$ kernel at $128 \times 128$, and four blocks of 256 channels and a $9 \times 9$ kernel at $64 \times 64$. The last two stages each begin with a stride-2 $3 \times 3$ convolution. Each block expands the channels twofold with a $1 \times 1$ convolution, applies the large depthwise kernel in parallel with a $5 \times 5$ one as in RepLKNet, projects back and adds a residual, with GELU activations~\cite{hendrycks2016gaussian} throughout.

The motivation was receptive field. At 0.4~m resolution a $13 \times 13$ kernel spans 5.2~m, a little more than the median car length of 4.61~m in our data, so one layer can see every pillar of a vehicle. A stack of $3 \times 3$ convolutions needs many more layers for comparable context, since the effective receptive field grows faster with kernel size than with depth~\cite{2022largekernel}.

The parallel branches support structural re-parameterisation~\cite{ding2021repvgg, 2022largekernel}, which folds the $5 \times 5$ branch and both BatchNorm layers into one large kernel at inference,
\begin{align*}
  W_{\text{fused}} & = \frac{\gamma_L}{\sqrt{\sigma_L^2+\varepsilon}} W_L + \text{pad}\left( \frac{\gamma_S}{\sqrt{\sigma_S^2+\varepsilon}} W_S \right) \\
  b_{\text{fused}} & = \left(\beta_L - \mu_L \frac{\gamma_L}{\sqrt{\sigma_L^2+\varepsilon}}\right)                                                      \\ &\quad + \left(\beta_S - \mu_S \frac{\gamma_S}{\sqrt{\sigma_S^2+\varepsilon}}\right)
\end{align*}
where $\gamma$, $\beta$, $\mu$ and $\sigma^2$ are the BatchNorm scale, shift, running mean and running variance of the large ($L$) and small ($S$) branches, unrelated to the encoder gate $\gamma$, the Smooth-L1 $\beta$ and the AdamW $\beta$.

The fusion is exact to single-precision round-off, with a maximum deviation of $1.1 \times 10^{-5}$ on outputs of magnitude 20 with TF32 disabled, and removes 78,848 parameters, 1.1\% of the model. It is not applied in any reported measurement, so all parameter counts and latencies in this paper are for the unfused model. Gradient checkpointing~\cite{chen2016training} is applied to the first two stages during training only, so training and inference memory are not directly comparable.

\subsection{Normal-guided feature modulation}
\label{subsec:ngfm}
\begin{figure*}[!t]
  \centering
  \includegraphics[width=\textwidth]{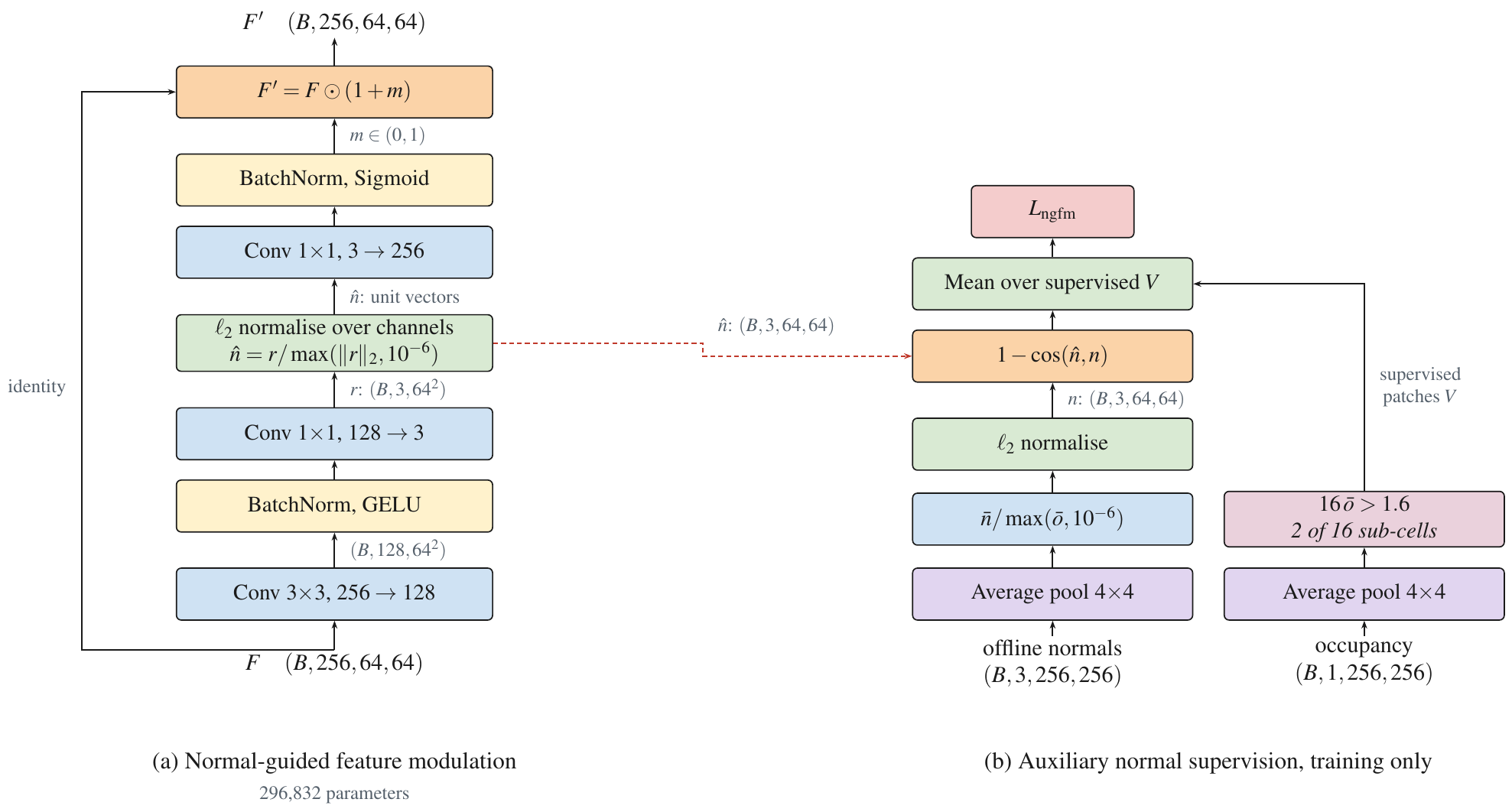}
  \caption{Normal-guided feature modulation at the $64\times64$ bottleneck. (a)~The forward branch, the only part used at inference. Because the gate $m$ is a sigmoid, $1+m$ lies in $(1,2)$, so a feature can be at most doubled and never suppressed, and the $3\rightarrow256$ expansion decides which orientations are amplified. (b)~The training-only target and loss. Offline normals are scattered to the $256\times256$ grid, pooled over $4\times4$ patches weighted by occupancy and renormalised, and a patch is supervised when at least two of its sixteen cells are occupied. A density weight in the implementation is identically one, so the loss is the unweighted mean shown.}
  \label{fig:ngfm}
\end{figure*}

At the $64 \times 64$ bottleneck, a branch predicts a unit surface normal per cell and uses it to gate the backbone features (Fig.~\ref{fig:ngfm}). Writing $F$ for the bottleneck feature map,

\begin{align}
  \label{eq:normal_features}
  h       & = \text{GELU}(\text{BN}(\text{Conv}_{3\times3}(F))), \quad r = \text{Conv}_{1\times1}(h) \\[1ex]
  \label{eq:normal_normalize}
  \hat{n} & = \frac{r}{\max(\Vert r \Vert_2, 10^{-6})}                                               \\[1ex]
  \label{eq:normal_modulate}
  m       & = \sigma(\text{BN}(\text{Conv}_{1\times1}(\hat{n}))), \quad F' = F \odot (1 + m)
\end{align}

Equations \eqref{eq:normal_features} and \eqref{eq:normal_normalize} predict a unit normal per cell, with the $3 \times 3$ convolution reducing 256 channels to 128, and \eqref{eq:normal_modulate} gates the features in the residual attention form of~\cite{wang2017residual}. Since $m \in (0, 1)$, the branch scales each feature by a factor between one and two and can never suppress one, and the $3 \rightarrow 256$ expansion decides which orientations are amplified.

The target comes from normals estimated offline by local plane fitting~\cite{Hoppe1992Surface, zhou2018open3d} on the full point cloud and oriented toward the sensor. They are scattered to the $256 \times 256$ grid, where a cell with at least three points takes the renormalised mean of their normals. Occupancy-weighted average pooling over $4 \times 4$ patches gives the $64 \times 64$ target, and a patch is supervised when at least two of its sixteen cells are occupied. Over the set $V$ of supervised patches the loss is

\begin{equation}
  \label{eq:ngfm_loss}
  L_{\text{ngfm}} = \frac{\sum_{v \in V} (1 - \cos(\hat{n}_v, n_v))}{\max(|V|, 1)}
\end{equation}

with weight $1.0$, added to the heatmap and box losses.

The implementation also weights each term of \eqref{eq:ngfm_loss} by a density map meant to favour densely observed patches. That map is derived by testing whether a cell of the encoder output is non-zero, and because the positional prior is added to every cell, the test always passes and the weight is identically one. The supervision mask comes from the true occupancy computed in the data pipeline, so the loss is correctly masked and, in effect, unweighted.

The intended mechanism was LUPI~\cite{VAPNIK2009544}. The normals are estimated within a $1$~m radius on the full cloud, whereas the network sees at most eight points per 0.4~m cell, so the target carries geometry that the input has partly lost. Section~\ref{subsec:aux} reports what the branch learned.

\subsection{Decoder and detection head}
\label{subsec:head}
The backbone output at stride 4 is restored to $256 \times 256$ by a two-stage decoder. Each stage doubles the resolution with a $2 \times 2$ transposed convolution of stride 2 and refines it with a $3 \times 3$ convolution, both followed by batch normalisation and GELU, and the first stage reduces 256 channels to 128. A stock \texttt{mmdet3d}~\cite{mmdet3d2020} CenterHead~\cite{yin2021center} then decodes at \texttt{out\_size\_factor = 1}, so every detection cell is one 0.4~m pillar. The decoder and the normal branch are independent. Run G keeps the decoder and drops the branch, and Section~\ref{subsec:decomposition} shows that head resolution, not the branch, accounts for most of the gap.

Head resolution matters because CenterHead assigns targets in head cells. It places a Gaussian at each ground-truth centre~\cite{zhou2019objects} with radius $\max(\text{min\_radius}, \lfloor\text{gaussian\_radius}(w, l)\rfloor)$, using the radius function of CornerNet~\cite{law2018cornernet}, $\text{min\_radius} = 2$ from CenterPoint~\cite{yin2021center} and a minimum overlap of 0.25, where the \texttt{mmdet3d} reference uses 0.1. Table~\ref{tab:radius_configurations} resolves this rule for the median CARLA car of $4.61 \times 1.97$~m.

\begin{table}[!t]
  \caption{Detection Head Radius Configurations}
  \label{tab:radius_configurations}
  \centering
  \setlength{\tabcolsep}{2.5pt}
  \begin{tabular}{l c c c c c}
    \hline
    Pillar, OS            & Head     & Car Extent          & Computed & After & Radius \\
    (runs)                & Cell (m) & (cells)             & Radius   & Clamp & (m)    \\
    \hline
    0.4~m, 4 (D, E)       & 1.60     & 2.88 $\times$ 1.23  & 0.90     & 2     & 3.20   \\
    0.2~m, 4 (C)          & 0.80     & 5.76 $\times$ 2.46  & 1.80     & 2     & 1.60   \\
    0.4~m, 1 (A, B, F, G) & 0.40     & 11.53 $\times$ 4.93 & 3.60     & 3     & 1.20   \\
    \hline
  \end{tabular}
\end{table}

At 1.6~m head cells, as in the budget-matched baseline, the computed radius of 0.90 cells floors to zero and is clamped to two cells, a Gaussian of radius 3.20~m around a car less than 2~m wide. The offset head~\cite{zhou2019objects} must also recover a quantisation error of up to one cell, or 1.6~m. At 0.4~m cells the radius is 1.20~m against a half-width near 1~m, and the offset head has at most 0.4~m to recover. The D$\to$F rung makes exactly this change and recovers 10.19 points (Section~\ref{subsec:decomposition}). We attribute that gain to the quantities that move with the head stride rather than to capacity, although the rung also adds 2.21~M parameters.

The regression loss shifts with the stride in the same way. The centre offset is expressed in head cells, and the \texttt{mmdet}~\cite{mmdetection} parameterisation of Smooth-L1~\cite{girshick2015fast} turns from quadratic to linear at $|x| = \beta$ in those units, so the physical transition is $\beta$ times the cell size. The runs use $\beta = 0.44$ at 0.4~m and 1.6~m cells and $\beta = 0.11$ at 0.8~m, which puts the transition at 17.6~cm, 70.4~cm and 8.8~cm. Against the median half-width of 0.99~m, the quadratic region covers 71\% of a car at 1.6~m cells but 8.9\% at 0.8~m. Changing the head stride without re-deriving these two constants therefore changes both the heatmap target and the regression objective.

Apart from input width, cell size and $\beta$, the head is identical in all seven runs. It uses \texttt{SeparateHead} with final kernel 3 and bias initialisation $-2.19$, the focal-loss prior initialisation of Lin et al.~\cite{lin2017focal} evaluated at $\pi \approx 0.1$ as in CenterPoint, a Gaussian focal classification loss~\cite{law2018cornernet, zhou2019objects}, Smooth-L1 regression at weight 0.5, log-space dimensions~\cite{yin2021center} and code weights $[0.5, 0.5, 1, 2, 2, 1, 2, 2, 0, 0]$. These regression settings differ from the \texttt{mmdet3d} CenterPoint reference~\cite{mmdet3d2020}, which uses L1 at weight 0.25 with code weights $[1, 1, 1, 1, 1, 1, 1, 1, 0.2, 0.2]$. The two zero weights mean that the velocity head is predicted in every run but never supervised, and no result depends on it. At test time every run decodes the 500 highest heatmap peaks, keeps those scoring at least 0.05 and applies rotated BEV non-maximum suppression at IoU 0.2, returning at most 150 boxes per scan.

\subsection{Scope of the Claims}
\label{subsec:scope}
Four claims motivated this design, and the evaluation supports none of them. They are listed here so that no later section is read as resting on one.
\begin{itemize}
  \item \textbf{Edge deployment.} GeoPillars was motivated by perception under sensor and compute limits, but it is the slowest model in the study at both measured batch sizes and the second most memory-intensive, behind the simpler baseline F (Section~\ref{subsec:cost}).

  \item \textbf{Intensity dependence.} We meant to measure how much the detector relies on intensity, but intensity is constant in every scan (Section~\ref{subsec:audit}), so runs B and E measure only insensitivity to an uninformative feature.

  \item \textbf{A 16,000-pillar constraint.} The regime was framed around \texttt{max\_voxels}, but no scan reaches that cap (Section~\ref{subsec:sparsity}). Only the eight-point cap binds.

  \item \textbf{Attribution of the architectural increment.} The F$\to$G rung makes six changes at once, so its 3.37 points cannot be credited to any single component (Section~\ref{subsec:decomposition}).
\end{itemize}

\section{Implementation}
\label{sec:implementation}
\subsection{Dataset collection}
\label{subsec:dataset}
We collected the dataset in CARLA 0.9.16~\cite{dosovitskiy2017carla} in synchronous mode at 20~Hz, with 100 warm-up ticks after each map load. The ego vehicle, a Lincoln MKZ 2020, carries one ray-cast LiDAR 1.84~m above its origin. The sensor is configured with 32 channels, 691,200 points per second (34,560 per revolution), a vertical field of view from $+10.67^\circ$ to $-30.67^\circ$, a 100~m range, a 10\% random point dropoff and 2~cm Gaussian range noise~\cite{carla_sensors}. CARLA applies the dropoff before ray casting and perturbs each point along its own ray, so angular positions are exact and only range is noisy~\cite{carla_sensors}. Points are converted to the nuScenes frame~\cite{caesar2020nuscenes} by swapping the CARLA $x$ and $y$ axes.

Collection covered four towns, three weather presets (ClearNoon, CloudyNoon, HardRainNoon) and five traffic profiles, from 8 to 15 vehicles with 30 pedestrians up to 50 to 70 vehicles with 180 pedestrians. Each scene draws one preset and one profile uniformly at random and is scheduled for 5,000 to 10,000 simulator ticks before the ego vehicle respawns at a new spawn point. A scene ends early if the ego vehicle gets stuck and is discarded if it crashes. After filtering, 154,206 scans remain, with 61,808 from Town03, 39,495 from Town05, 35,823 from Town10HD\_Opt and 17,080 from Town02, split into 123,364 for training and 30,842 for validation.

Annotations follow the nuScenes taxonomy~\cite{caesar2020nuscenes} and contain 1,562,560 car, 229,846 pedestrian, 116,261 bus and 106,551 truck boxes. The car class comes with three caveats. It is a catch-all, because every CARLA vehicle model that is not a truck, bus, motorcycle or bicycle is labelled as a car, which includes 104,872 ambulance and 91,480 police-vehicle boxes. It includes 101,349 static map vehicles, 6.5\% of the class, taken from CARLA's level bounding boxes rather than from spawned actors. And 44.5\% of validation car yaws lie within $10^{-6}$~rad of $\pm\pi/2$, because waypoint-following traffic shares lane headings with the ego vehicle and the collector's yaw convention maps a parallel heading to exactly $+\pi/2$. Heading in this dataset is therefore close to a binary decision (Section~\ref{subsec:orientation}).

Ground truth is the offline filter, cars within 50~m with at least five LiDAR points, and the training-time filters remove nothing further from the validation split. That leaves 313,179 scored car boxes with a median of 36 points and a median range of 24.1~m, of which 130,048 lie within 20~m, 132,628 between 20 and 40~m and 50,503 beyond. The median footprint is $4.61 \times 1.97$~m with an interquartile range of $0.99 \times 0.33$~m, and 4.0\% of validation scans contain no scored car.

Pedestrians, buses and trucks stay in the point cloud with their boxes removed, so detections on them count as false positives. The validation split has 90,357 such boxes in 26,657 frames. Counting a false positive as a hit when it overlaps a non-car box at BEV IoU 0.3 or more, these objects account for 5.16\%, 5.89\% and 2.65\% of the IoU-0.5 false positives of runs A, F and D. Trucks make up 83\% to 86\% of the hits. Pedestrians cannot be hit at all, because a car-sized box cannot reach IoU 0.3 against a 0.7~m walker. The effect makes absolute AP slightly pessimistic and differs between runs, but at a few percent of false positives it cannot explain any gap in Section~\ref{sec:results}.

Normals were estimated offline with Open3D~\cite{zhou2018open3d} using a hybrid KD-tree search with a 1~m radius and at most 30 neighbours, oriented toward the sensor and stored as columns 4 to 6 of each point. These columns are removed before voxelisation, so the normals reach the network only as the auxiliary target. They were computed on every point before any class filtering, which matters for the auxiliary loss (Section~\ref{subsec:audit}).

The 80/20 split is a seeded shuffle of frames, not of scenes. As a result, 95.2\% of validation frames have the previous or next saved frame of the same town, 0.5~s away in simulator time, in the training set, and 61.5\% have both. This is the non-independence between training and test samples that makes held-out estimates optimistic~\cite{KAPOOR2023100804}. There is no scene-disjoint set, and since no run used \texttt{save\_best}, checkpoints were chosen by epoch count alone.

\subsection{The sparsity regime}
\label{subsec:sparsity}
The configuration sets \texttt{max\_num\_points = 8} and \texttt{max\_voxels = 16000} (40,000 at test), and only the first binds. The densest of the 154,206 scans occupies 5,085 pillars and the 99th percentile of the training split is 4,476, more than a factor of three below the pillar cap. The point cap, by contrast, fills all eight slots in 25.2\% of occupied pillars and drops points from 21.7\% of them. Across the training split it discards 46.6\% of in-range points, from scans that hold on average 26,147 in-range points in 3,320 occupied pillars.

Neither cap comes from a sensor specification, and we do not claim that this configuration matches a deployed system. It is a controlled degradation of a 32-channel input, similar in spirit to the beam downsampling that LiDAR Distillation~\cite{wei2022lidar} uses to align point density between sensors, but it caps points per pillar rather than removing beams. Every run except C is trained and evaluated under it.

\subsection{Implementation Audit}
\label{subsec:audit}
A post-training audit of the code and the checkpoints found eight properties that affect how Section~\ref{sec:results} should be read. Each is listed with its cause and, where measurable, its cost.
\begin{enumerate}
  \item \emph{Intensity is constant.} The collection script sets \texttt{atmosphere\_attenuation\_rate} to zero, and CARLA computes intensity as $I_0 \exp(-a \cdot d)$ for attenuation $a$ and distance $d$~\cite{carla_sensors}, so every point in all 154,206 scans has intensity 1.0. The weather presets do not change sensor attributes. Every model therefore learned from geometry alone, and every run except B and E carries one constant input feature.

  \item \emph{The auxiliary target is mostly background.} Normals were computed before class filtering, so the auxiliary task supervises the whole scene while the detector learns only cars. Over 500 training scans, 8.79\% of bottleneck cells are supervised. Of those patches, 6.38\% contain car points, 2.18\% contain other objects but no car, and 91.44\% contain no object at all, with each share moving by less than 0.1 points under a second augmentation draw. Because the loss is an unweighted mean (item 6), these are also the shares of the loss weight. The auxiliary tasks of SA-SSD~\cite{He2020_SASSD}, foreground segmentation and centre estimation, are defined by the object boxes, and ours is not.

  \item \emph{The regression loss is not on a common footing.} The Smooth-L1~\cite{girshick2015fast} $\beta$ puts the quadratic-to-linear transition at 17.6~cm, 70.4~cm or 8.8~cm depending on the head cell (Section~\ref{subsec:head}).

  \item \emph{The validation split is not scene-disjoint} (Section~\ref{subsec:dataset}).

  \item \emph{A learned positional bias holds most of the parameters} (Section~\ref{subsec:prior}) and is worth 0.82 AP points at inference (Section~\ref{subsec:prior_ablation}).

  \item \emph{The density weight in the auxiliary loss is identically one} (Section~\ref{subsec:ngfm}). The supervision mask itself is correct.

  \item \emph{Two metric definitions were defective.} The original mean BEV IoU counted matches that survived IoU 0.7 at both thresholds and was 1.13 to 1.66 points optimistic. The original orientation error followed nuScenes~\cite{caesar2020nuscenes} and charged $\pi$ for a box yawed by $\pi$, which is the same cuboid reversed. The corrected evaluator restricts the first to one threshold and reports the second alongside a folded error and a heading-flip rate. Neither change affects average precision, and the other 25 of the 26 original metrics are unchanged. We omit mean BEV IoU entirely, because it averages over matched detections only and so rewards a model that matches fewer, easier boxes.

  \item \emph{Training and inference keep different points.} Hard voxelisation~\cite{pmlr-v100-zhou20a} in mmcv~\cite{mmcv} keeps the first points of each pillar in arrival order, and \texttt{PointShuffle}~\cite{mmdet3d2020} runs last in every training pipeline and in no test pipeline. Over 200 validation scans and five shuffle draws, 98.2\% of overflowing pillars keep a different set, with a mean overlap of 4.37 of 8 against 4.36 for independent uniform draws. The unshuffled set also sits 19~cm higher on average, mostly in the pillars with the largest height range. Inserting the shuffle at test time moves the AP@0.7 of run A from 0.8133 to 0.8136. The encoder is invariant to point order, so the 0.03-point change reflects only which eight points survive.
\end{enumerate}

\subsection{Training protocol}
\label{subsec:protocol}
All seven models were trained for eight epochs at a global batch of 112 on the full training split. Optimisation used AdamW~\cite{loshchilov2019adamw} at $\beta = (0.95, 0.99)$ and weight decay 0.01, a one-cycle schedule~\cite{smith2019superconvergence} peaking at $2 \times 10^{-4}$ after 10\% warmup, gradient-norm clipping~\cite{pascanu2013difficulty} at 35 and \texttt{bfloat16} mixed precision~\cite{micikevicius2018mixed, kalamkar2019bfloat16}. The schedule was configured for 8,820 steps, of which 8,816 ran. Augmentation is identical across runs and consists of global rotation by $\pm 22.5^\circ$ and scaling by $\pm 5\%$, both taken from the CenterPoint nuScenes configuration~\cite{yin2021center, mmdet3d2020}, and a horizontal flip as in PointPillars~\cite{lang2019pointpillars}. No run used ground-truth pasting~\cite{yan2018second} or per-box noise~\cite{2018voxelnet}.

The runs did not share hardware. A, B and D trained on an H100 PCIe, C and F on an A100-SXM4-40GB, E on an A10, and G on two RTX 5090s under distributed data parallel~\cite{li2020pytorchdistributed} with 56 samples per GPU. Training memory and wall-clock time are therefore not comparable across runs, and only the former is reported, in Table~\ref{tab:model_zoo}, with that caveat. No configuration requested synchronised batch normalisation, so G computed its statistics over 56 samples per rank instead of 112, one of the six changes bundled into the F$\to$G rung.

\begin{figure}[!t]
  \centering
  \includegraphics[width=\columnwidth]{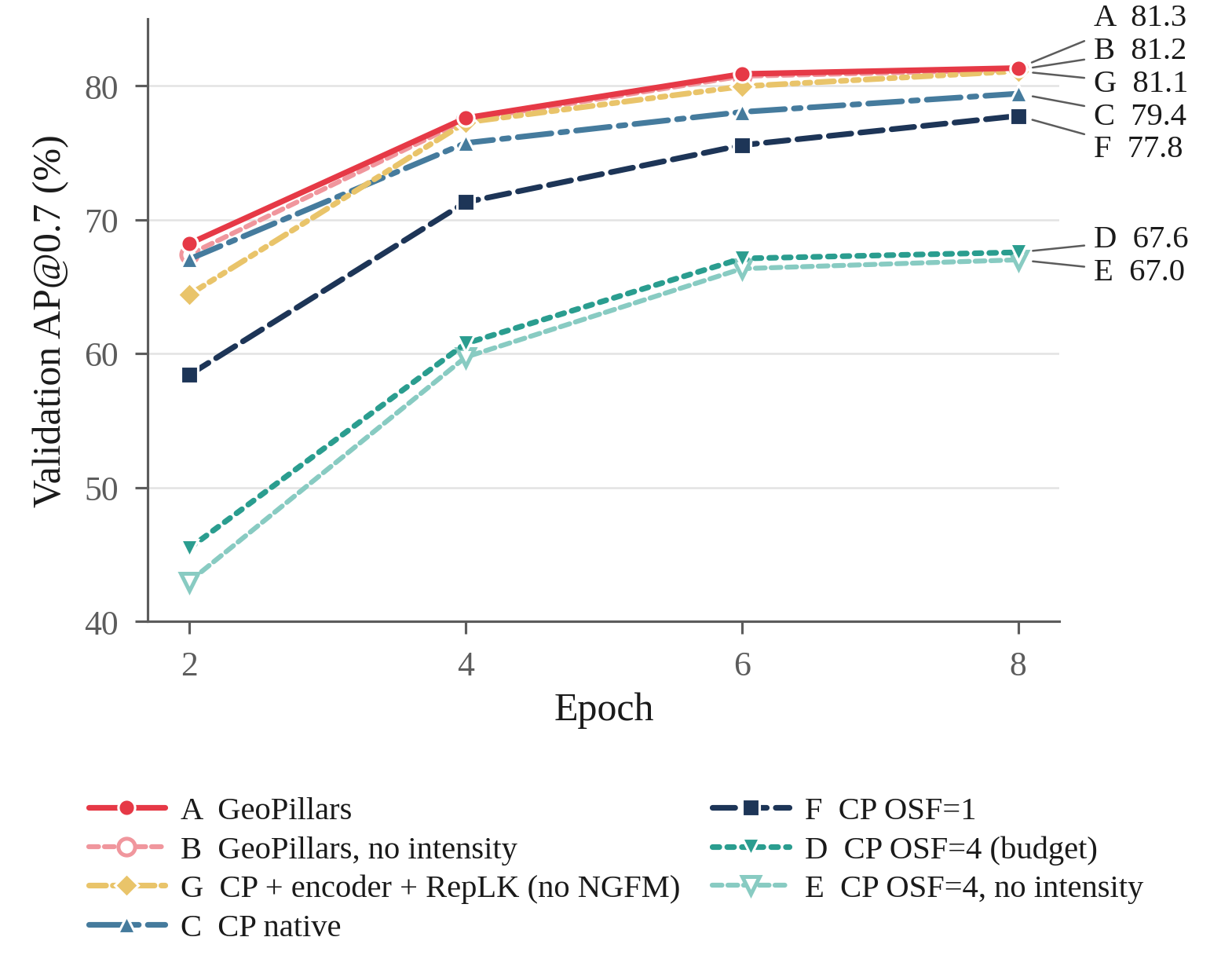}
  \caption{Validation AP@0.7 at epochs 2, 4, 6 and 8, as logged during training. The epoch-8 points of B and G lie 0.01 and 0.03 points below Table~\ref{tab:main}, within the re-evaluation tolerance. Every curve is still rising, so the epoch-8 ordering need not hold at convergence.}
  \label{fig:val_ap}
\end{figure}

Each run used a single logged seed and is not exactly reproducible, because cuDNN benchmarking was enabled and deterministic mode was not. Every validation curve was still rising at epoch eight (Fig.~\ref{fig:val_ap}), so eight epochs marks where training stopped, not where it converged.

All models are built on mmdetection3d 1.4.0~\cite{mmdet3d2020}, mmcv 2.1.0~\cite{mmcv}, mmdet 3.3.0~\cite{mmdetection} and mmengine 0.10.7~\cite{mmengine2022}, vendored with small compatibility patches for recent PyTorch releases and for the configuration format of the CenterPoint runs. No detection operator, loss, data transform or metric was modified. GeoPillars supplies its own encoder, backbone, NGFM, upsampler, data transforms and evaluator and uses the stock CenterHead and training loop. Runs A and B wrap these modules in a detector class that adds the auxiliary loss, and run G places the same encoder, backbone and upsampler in the stock CenterPoint detector through three adapter modules. Training used PyTorch 2.9.0~\cite{paszke2019pytorch} with CUDA 13.0, and the re-evaluations in Section~\ref{sec:results} used PyTorch 2.7.1 with CUDA 12.8, reproducing the training-time epoch-8 AP within 0.03 points for every run.

\section{Results}
\label{sec:results}
Apart from the training curves in Fig.~\ref{fig:val_ap}, every number in this section comes from re-evaluating the epoch-8 checkpoints on the CARLA validation split or on the zero-shot nuScenes set. The CARLA split is the one monitored during training and is not scene-disjoint from the training data. Average precision is 40-point interpolated~\cite{simonelli2019disentangling} over exact rotated BEV IoU.

\subsection{Main Comparison}
\label{subsec:main}
Table~\ref{tab:main} reports all seven runs. GeoPillars reaches 0.8133 AP@0.7, 1.91 points above CenterPoint at its native budget (C, 0.7942) and 13.76 points above the same architecture at the GeoPillars budget (D, 0.6757).

Most of the separation is precision. At IoU 0.5 the seven runs find between 295,647 and 302,316 of the 313,179 cars, a spread of 2.2\%, while their false-positive counts differ by a factor of 2.8. D, for example, emits 1,081,675 detections against 591,580 for A and finds 2\% fewer cars. At IoU 0.7 the runs also separate in recall. The true-positive spread widens to 10.6\%, D's ranked list ends at 0.773 recall against 0.858 for A, and at a recall of 0.75, which every run reaches, precision is 0.7628 for D and 0.9057 for A.

\begin{table*}[!t]
  \caption{CARLA validation split, epoch 8, cars. AP is 40-point interpolated over exact rotated BEV IoU. Orientation is folded into $[0, \pi/2]$ over matches at IoU $\ge 0.5$, and the flip rate is the share of those matches whose unfolded error exceeds $\pi/2$. Detections, false-positive share and precision at recall 0.75 refer to the full ranked list at IoU 0.7. Mean AP is omitted because with one class it is the mean of the two AP columns, and mean BEV IoU for the reason in Section~\ref{subsec:audit}.}
  \label{tab:main}
  \centering
  \begin{tabular}{c l c c c c c c c r r c}
    \hline
        &                          & \multicolumn{2}{c}{Overall} & \multicolumn{3}{c}{AP@0.7 by range (m)} & \multicolumn{2}{c}{Orientation} & \multicolumn{3}{c}{At IoU 0.7}                                                                                \\
    \cline{3-4}\cline{5-7}\cline{8-9}\cline{10-12}
    Run & Model                    & AP@0.7                      & AP@0.5                                  & 0--20                           & 20--40                         & $\ge$40 & mAOE$_\mathrm{f}$ & Flip   & Det.      & FP     & $P$ @ $r{=}0.75$ \\
    \hline
    A   & GeoPillars               & \textbf{0.8133}             & 0.9315                                  & 0.9046                          & 0.7603                         & 0.6191  & 0.0469            & 0.80\% & 591,580   & 54.6\% & 0.9057           \\
    B   & GeoPillars, xyz          & 0.8121                      & 0.9306                                  & 0.9007                          & 0.7460                         & 0.6186  & 0.0473            & 0.96\% & 590,656   & 54.5\% & 0.9020           \\
    C   & CenterPoint, native      & 0.7942                      & 0.9192                                  & 0.8926                          & 0.7231                         & 0.6052  & 0.0416            & 3.66\% & 873,917   & 69.2\% & 0.8835           \\
    D   & CenterPoint, budget      & 0.6757                      & 0.8955                                  & 0.8009                          & 0.5967                         & 0.4469  & 0.0481            & 0.92\% & 1,081,675 & 77.6\% & 0.7628           \\
    E   & CenterPoint, budget, xyz & 0.6701                      & 0.8940                                  & 0.7955                          & 0.5921                         & 0.4430  & 0.0505            & 0.99\% & 1,108,879 & 78.3\% & 0.7472           \\
    F   & CenterPoint, OS${=}$1    & 0.7776                      & 0.9247                                  & 0.8772                          & 0.7089                         & 0.5700  & 0.0517            & 1.55\% & 680,132   & 61.8\% & 0.8669           \\
    G   & GeoPillars without NGFM  & 0.8113                      & 0.9312                                  & 0.8848                          & 0.7444                         & 0.6152  & 0.0463            & 0.86\% & 597,194   & 55.2\% & 0.9005           \\
    \hline
  \end{tabular}
\end{table*}

\subsection{Decomposition}
\label{subsec:decomposition}
\begin{figure}[!t]
  \centering
  \includegraphics[width=\columnwidth]{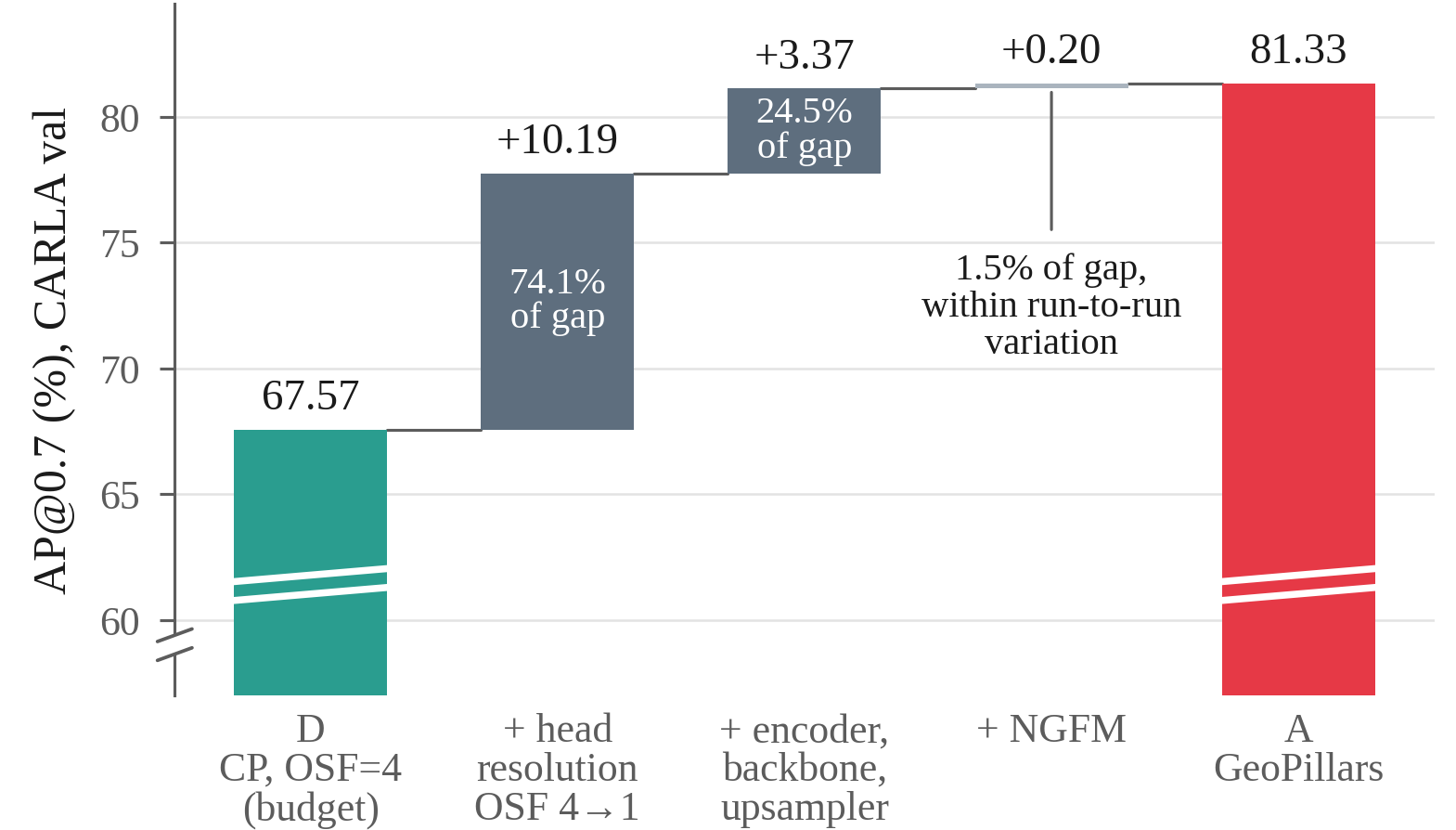}
  \caption{Decomposition of the 13.76-point AP@0.7 gap between the budget-matched CenterPoint baseline (D) and GeoPillars (A). Raising the head resolution (D$\to$F) recovers 10.19 points. The F$\to$G rung makes six simultaneous changes and adds 3.37 points that cannot be credited to any one of them, and the normal branch (G$\to$A) adds 0.20 points, within run-to-run variation. The vertical axis is broken.}
  \label{fig:decomposition}
\end{figure}

Table~\ref{tab:decomposition} and Fig.~\ref{fig:decomposition} split the 13.76-point gap along the ladder D$\to$F$\to$G$\to$A, in which each rung is the change from one run to the next. All four runs are re-evaluated on the same split, so the three increments, 10.19, 3.37 and 0.20 points, sum exactly to the gap.

The D$\to$F rung raises the head from \texttt{out\_size\_factor} 4 to 1, changes only the neck strides this requires, and recovers 10.19 points, 74\% of the gap. F also trained on an A100 rather than an H100 and under a different seed. Two constants move with the stride without appearing in the model description. The Gaussian radius~\cite{law2018cornernet, yin2021center} clamps to a region wider than the car at 1.6~m cells and becomes proportionate to it at 0.4~m cells. The Smooth-L1 transition~\cite{girshick2015fast} narrows from 70.4~cm to 17.6~cm of displacement even though both runs set $\beta = 0.44$ (Section~\ref{subsec:head}). The 10.19 points therefore bundle a finer output grid, a better-matched target assignment and a regression loss that tightens around the centre.

The F$\to$G rung adds 3.37 points but is not a controlled comparison. It swaps the encoder, backbone and neck, narrows the head input from 384 to 128 channels, adds the positional prior and, because G alone trained on two GPUs, changes the batch-normalisation statistics. F and A differ by only 3.2\% in parameters, 7.05~M against 7.28~M, so the 3.57 points that F concedes to A are not bought with capacity. But 4.19~M of G's 6.98~M parameters are the positional prior, so net of it the rung trades about 4.3~M parameters of convolutions for a lookup table indexed by position.

Evaluation noise is bounded by resampling the validation frames 500 times with replacement. A detection's greedy match depends only on higher-scoring detections in its own frame, so average precision is exact on any resample and no re-matching is needed. The bootstrap standard deviation of AP@0.7 is 0.07 to 0.09 points for six runs and 0.34 for D, whose precision-recall curve is the least stable. The 10.19 and 3.37-point increments exceed this by a factor of thirty or more and also clear the 0.13 to 0.57 points of run-to-run variation in Section~\ref{subsec:inputdim}. The 0.20-point increment does not, and Section~\ref{subsec:aux} examines it.

\begin{table}[!t]
  \caption{Decomposition of the 13.76-point AP@0.7 gap. Each row is a run of Table~\ref{tab:model_zoo}, and $\Delta$ is the change from the row above. The increments sum exactly to the gap and account for 74.1\%, 24.5\% and 1.5\% of it. D$\to$F changes only the head stride and the neck strides it requires, while F$\to$G changes six things at once. Parameters in parentheses exclude the positional prior.}
  \label{tab:decomposition}
  \centering
  \setlength{\tabcolsep}{4pt}
  \begin{tabular}{l c c l}
    \hline
    Run                                    & AP@0.7 & $\Delta$ & Params          \\
    \hline
    D \enspace CenterPoint, budget         & 67.57  & --       & 4.84~M          \\
    F \enspace $+$ head resolution         & 77.76  & $+$10.19 & 7.05~M          \\
    G \enspace $+$ encoder, backbone, neck & 81.13  & $+$3.37  & 6.98~M (2.78~M) \\
    A \enspace $+$ NGFM branch             & 81.33  & $+$0.20  & 7.28~M (3.08~M) \\
    \hline
  \end{tabular}
\end{table}

\subsection{Orientation at Range}
\label{subsec:orientation}
\begin{figure}[!t]
  \centering
  \includegraphics[width=\columnwidth]{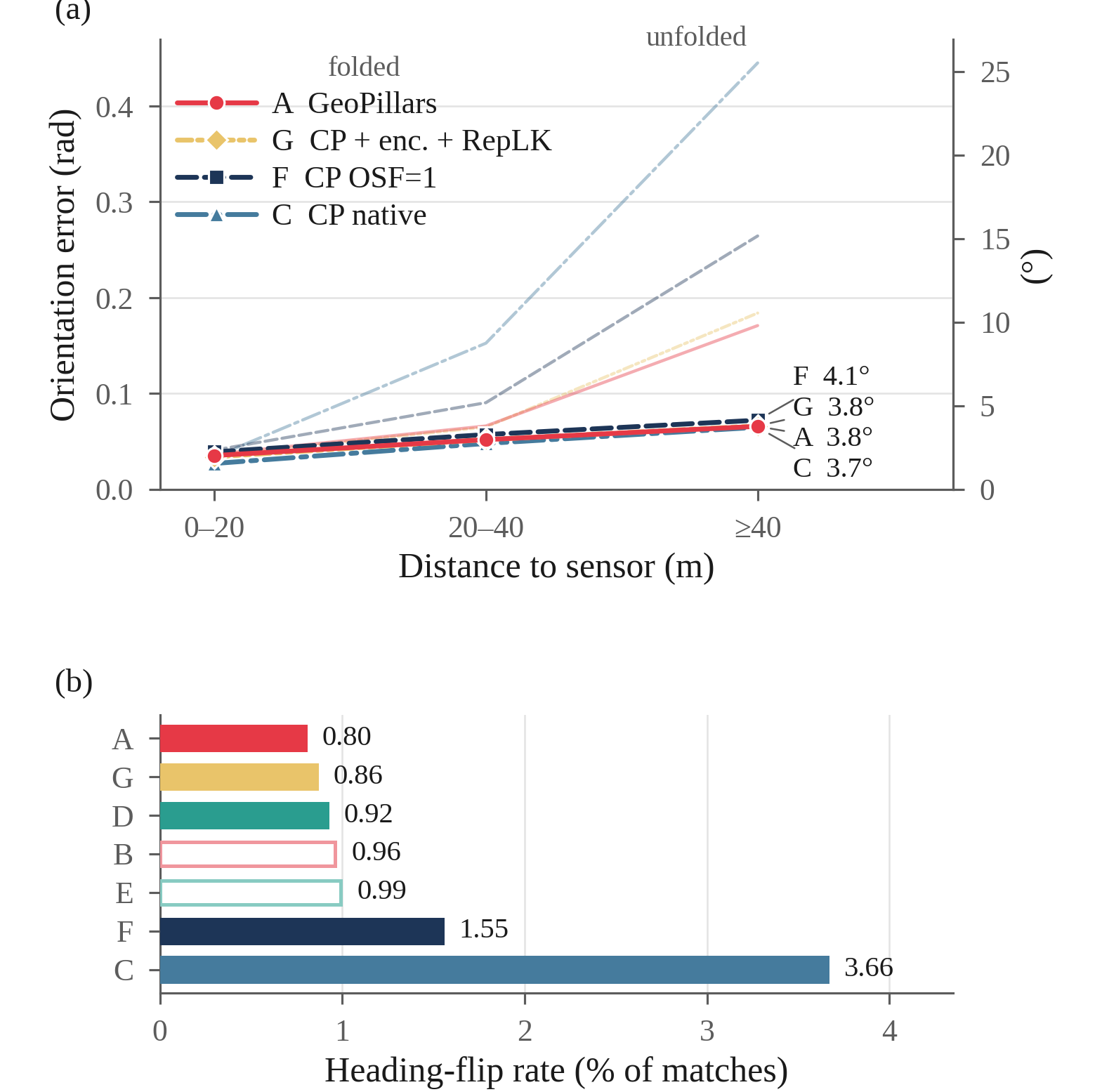}
  \caption{Orientation on true positives at BEV IoU $\ge 0.5$. (a) Mean orientation error against range for A, G, F and C, wrapped into $[0, \pi]$ (thin lines) and folded into $[0, \pi/2]$ (thick lines). The unfolded curves fan out with range, with C reaching 0.445~rad beyond 40~m, while the folded curves converge to 0.065 to 0.072~rad. (b) Heading-flip rate for all seven runs. C uses a different Smooth-L1 $\beta$ (0.11 against 0.44).}
  \label{fig:maoe_distance}
\end{figure}

Orientation is where the choice of metric decides the winner (Table~\ref{tab:orientation}, Fig.~\ref{fig:maoe_distance}). The originally reported mAOE wraps the yaw error into $[0, \pi]$~\cite{caesar2020nuscenes} and so charges $\pi$ for a reversed heading. Under it, C is the worst of the seven runs at 0.1489~rad and A the best at 0.0700. Folding heading reversal out, into $[0, \pi/2]$, makes C the best at 0.0416~rad and F the worst at 0.0517, and narrows the spread of all seven runs to $2.4^\circ$ to $3.0^\circ$. The change is largest beyond 40~m, where C's unfolded error of 0.4451~rad becomes a folded error of 0.0646, against 0.0656 for A. The 0.27~rad long-range gap between C and A in the unfolded metric is therefore entirely heading reversal.

What separates the models is how often they reverse a heading. A does so in 0.80\% of matches and C in 3.66\%, 4.6 times as often. The difference survives the correction, but this dataset tests it weakly, because 44.5\% of validation yaws sit at $\pm\pi/2$ and the flip rate is close to a classification error on a near-binary target. C also uses a different Smooth-L1 $\beta$ from the other runs (Section~\ref{subsec:audit}). The normal branch does not help orientation either. A and G differ by 0.0006~rad in folded error, in G's favour, and by 0.06 points of flip rate, in A's.

\begin{table}[!t]
  \caption{Orientation on true positives at BEV IoU $\ge 0.5$. mAOE wraps the yaw error into $[0, \pi]$, as originally reported, and mAOE$_\mathrm{f}$ also folds heading reversal out, into $[0, \pi/2]$. The flip rate is the share of matches whose unfolded error exceeds $\pi/2$. C ranks worst under mAOE and best under mAOE$_\mathrm{f}$.}
  \label{tab:orientation}
  \centering
  \begin{tabular}{c c c c c c c}
    \hline
        & \multicolumn{2}{c}{Overall (rad)} &                   & \multicolumn{3}{c}{mAOE$_\mathrm{f}$ by range (m)}                             \\
    \cline{2-3}\cline{5-7}
    Run & mAOE                              & mAOE$_\mathrm{f}$ & Flip rate                                          & 0--20  & 20--40 & $\ge$40 \\
    \hline
    A   & 0.0700                            & 0.0469            & 0.80\%                                             & 0.0351 & 0.0517 & 0.0656  \\
    B   & 0.0748                            & 0.0473            & 0.96\%                                             & 0.0355 & 0.0525 & 0.0650  \\
    C   & 0.1489                            & 0.0416            & 3.66\%                                             & 0.0268 & 0.0477 & 0.0646  \\
    D   & 0.0742                            & 0.0481            & 0.92\%                                             & 0.0338 & 0.0532 & 0.0737  \\
    E   & 0.0786                            & 0.0505            & 0.99\%                                             & 0.0364 & 0.0556 & 0.0751  \\
    F   & 0.0966                            & 0.0517            & 1.55\%                                             & 0.0389 & 0.0570 & 0.0720  \\
    G   & 0.0710                            & 0.0463            & 0.86\%                                             & 0.0337 & 0.0517 & 0.0660  \\
    \hline
  \end{tabular}
\end{table}

\subsection{Auxiliary Supervision}
\label{subsec:aux}
Run G is run A without the normal-guided branch. The two checkpoints differ by exactly 296,832 parameters, all in the seven weight tensors of the branch, and the 311 tensors they share have identical shapes. The comparison is clean at the level of weights but not of training, since G ran on two GPUs with per-rank batch statistics, on different hardware and with a different seed.

The branch is worth 0.20 points between the rounded APs of Table~\ref{tab:main} and 0.208 points between the unrounded ones. A paired frame-level bootstrap gives a 95\% interval of $[+0.14, +0.29]$, with all 500 resamples agreeing in sign, so the difference is not an artefact of the finite validation set. It is, however, within the 0.13 to 0.57 points of run-to-run variation in Section~\ref{subsec:inputdim}, and we do not claim it as an effect.

\begin{figure}[!t]
  \centering
  \includegraphics[width=\columnwidth]{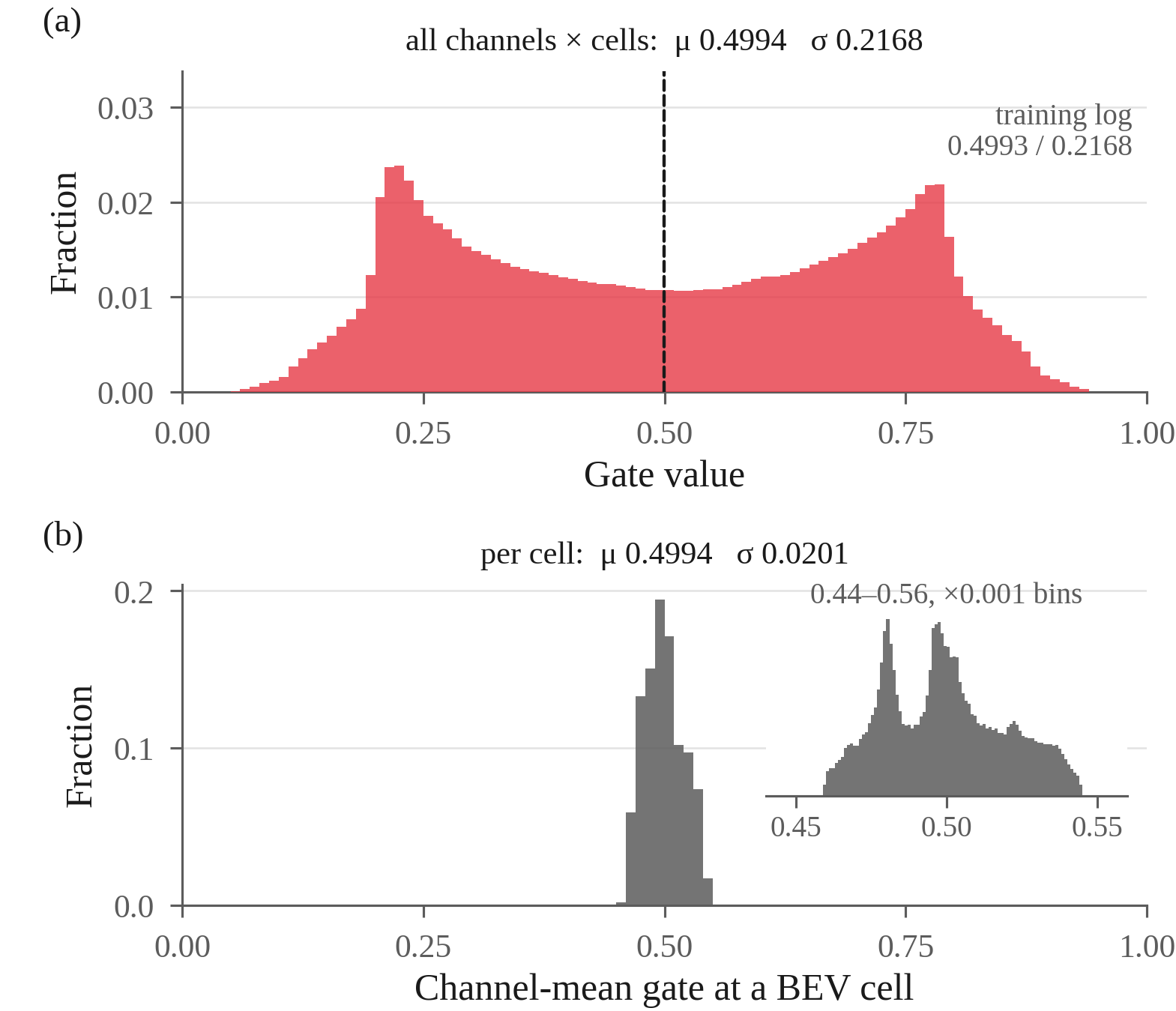}
  \caption{The normal-guided gate of run A at epoch 8, over 64 validation scans. (a) Gate values pooled over channels and cells, whose mean and standard deviation lie within 0.0002 of the training log. (b) The same values averaged over channels at each BEV cell, which collapse to $0.4994 \pm 0.0201$.}
  \label{fig:ngfm_mask}
\end{figure}

Fig.~\ref{fig:ngfm_mask} shows what the gate learned. Pooled over channels and cells it is broad and unsaturated, with mean 0.4994 and standard deviation 0.2168, matching the training log to three decimal places. Averaged over channels, however, it collapses to $0.4994 \pm 0.0201$ at every BEV cell, so the net gain $1 + m$ is about 1.50 everywhere. A $1 \times 1$ convolution of a unit normal has only two degrees of freedom per cell, and the branch uses them to reweight channels rather than to select regions.

\subsection{Input Dimensionality}
\label{subsec:inputdim}
Removing the intensity feature costs 0.13 points for GeoPillars (A 0.81335, B 0.81206) and 0.57 for the budget-matched CenterPoint (D 0.67575, E 0.67010). Because the feature is constant in every scan, neither number measures intensity dependence.

The two pairs instead measure run-to-run variation. Each pair trains the same architecture on the same data and schedule and differs in seed and in one constant input feature, and D and E also differ in GPU type and data-loader settings. Both differences exceed their evaluation noise, and both paired bootstrap intervals exclude zero. With no seed repeats in the study, these near-replicates give the only estimate of run-to-run variation, 0.13 to 0.57 points, and no difference of this size or smaller, including the 0.20 points of the normal branch, can be separated from the seed.

\subsection{Zero-Shot nuScenes}
\label{subsec:zeroshot}
Table~\ref{tab:zeroshot} and Fig.~\ref{fig:sim2real} report transfer to 6,743 real nuScenes keyframes without adaptation, under the protocol of Appendix~\ref{app:zeroshot}. No model transfers. Retention ranges from 3.67\% (E) to 12.20\% (C), and the CARLA ordering does not survive, since the native-budget baseline transfers best in both absolute AP and retained fraction.

\begin{figure}[!t]
  \centering
  \includegraphics[width=\columnwidth]{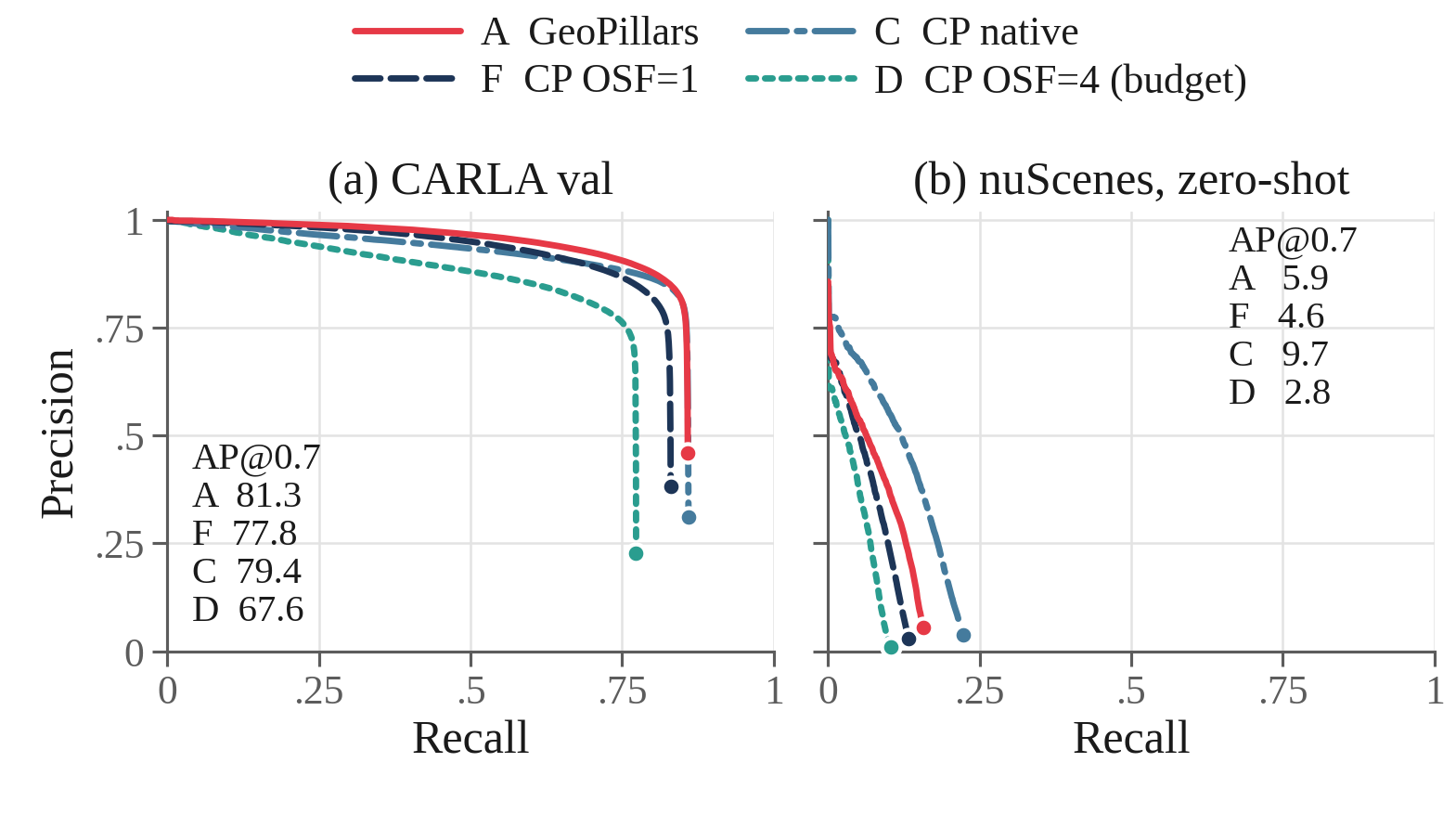}
  \caption{Precision and recall at BEV IoU 0.7 for runs A, C, D and F on (a) the CARLA validation split and (b) nuScenes zero-shot. On nuScenes no model exceeds 0.25 recall.}
  \label{fig:pr_curves}
\end{figure}

Recall is the main failure. At IoU 0.7 no ranked list passes 0.223 recall, and at IoU 0.5 the best reaches 0.413, so at least 58.7\% of the scored cars are never matched at any score. Precision at 5\% recall is 0.50 to 0.67 for A, B, C, F and G and 0.34 to 0.38 for D and E (Fig.~\ref{fig:pr_curves}), so the models still recognise cars but find far too few of them. Localisation degrades less. The share of IoU-0.5 matches that also pass IoU 0.7 falls from 0.81 to 0.89 on CARLA to 0.35 to 0.54 on nuScenes.

On nuScenes only the unfolded orientation error is reported, and its values near 1.0~rad are not interpreted.

A simple comparison does not support the absolute positional prior as the cause of the collapse. Of the four full-resolution runs, the three that carry the prior retain 6.93\% to 8.18\% and F, which has none, retains 5.86\%. Those three also differ from F in encoder, backbone, neck and head width, so the comparison settles nothing, and the decisive experiment of zeroing the prior on nuScenes was not run.

\begin{table*}[!t]
  \caption{Zero-shot transfer to 6,743 nuScenes keyframes with 42,705 scored car boxes and intensity overwritten to 1.0. No model saw nuScenes in training. Retained is nuScenes AP@0.7 as a percentage of the model's CARLA AP@0.7, and $r_\mathrm{max}$ is the highest recall the ranked list reaches. mAOE is the unfolded error.}
  \label{tab:zeroshot}
  \centering
  \begin{tabular}{c l c c c c c c c r}
    \hline
        &                          & \multicolumn{2}{c}{AP@0.7} &          & \multicolumn{2}{c}{nuScenes} & \multicolumn{2}{c}{$r_\mathrm{max}$} &                                      \\
    \cline{3-4}\cline{6-7}\cline{8-9}
    Run & Model                    & CARLA                      & nuScenes & Retained                     & AP@0.5                               & mAOE   & IoU 0.5 & IoU 0.7 & Det.    \\
    \hline
    A   & GeoPillars               & 0.8133                     & 0.0592   & 7.28\%                       & 0.1823                               & 0.9964 & 0.340   & 0.157   & 124,722 \\
    B   & GeoPillars, xyz          & 0.8121                     & 0.0664   & 8.18\%                       & 0.1862                               & 1.0822 & 0.355   & 0.173   & 124,434 \\
    C   & CenterPoint, native      & 0.7942                     & 0.0969   & 12.20\%                      & 0.2070                               & 0.8429 & 0.413   & 0.223   & 255,230 \\
    D   & CenterPoint, budget      & 0.6757                     & 0.0283   & 4.19\%                       & 0.1228                               & 1.0099 & 0.285   & 0.104   & 429,495 \\
    E   & CenterPoint, budget, xyz & 0.6701                     & 0.0246   & 3.67\%                       & 0.1190                               & 1.0133 & 0.279   & 0.098   & 413,300 \\
    F   & CenterPoint, OS${=}$1    & 0.7776                     & 0.0456   & 5.86\%                       & 0.1511                               & 1.0045 & 0.314   & 0.133   & 190,514 \\
    G   & GeoPillars without NGFM  & 0.8113                     & 0.0562   & 6.93\%                       & 0.1619                               & 0.9969 & 0.331   & 0.155   & 136,611 \\
    \hline
  \end{tabular}
\end{table*}
\begin{figure}[!t]
  \centering
  \includegraphics[width=\columnwidth]{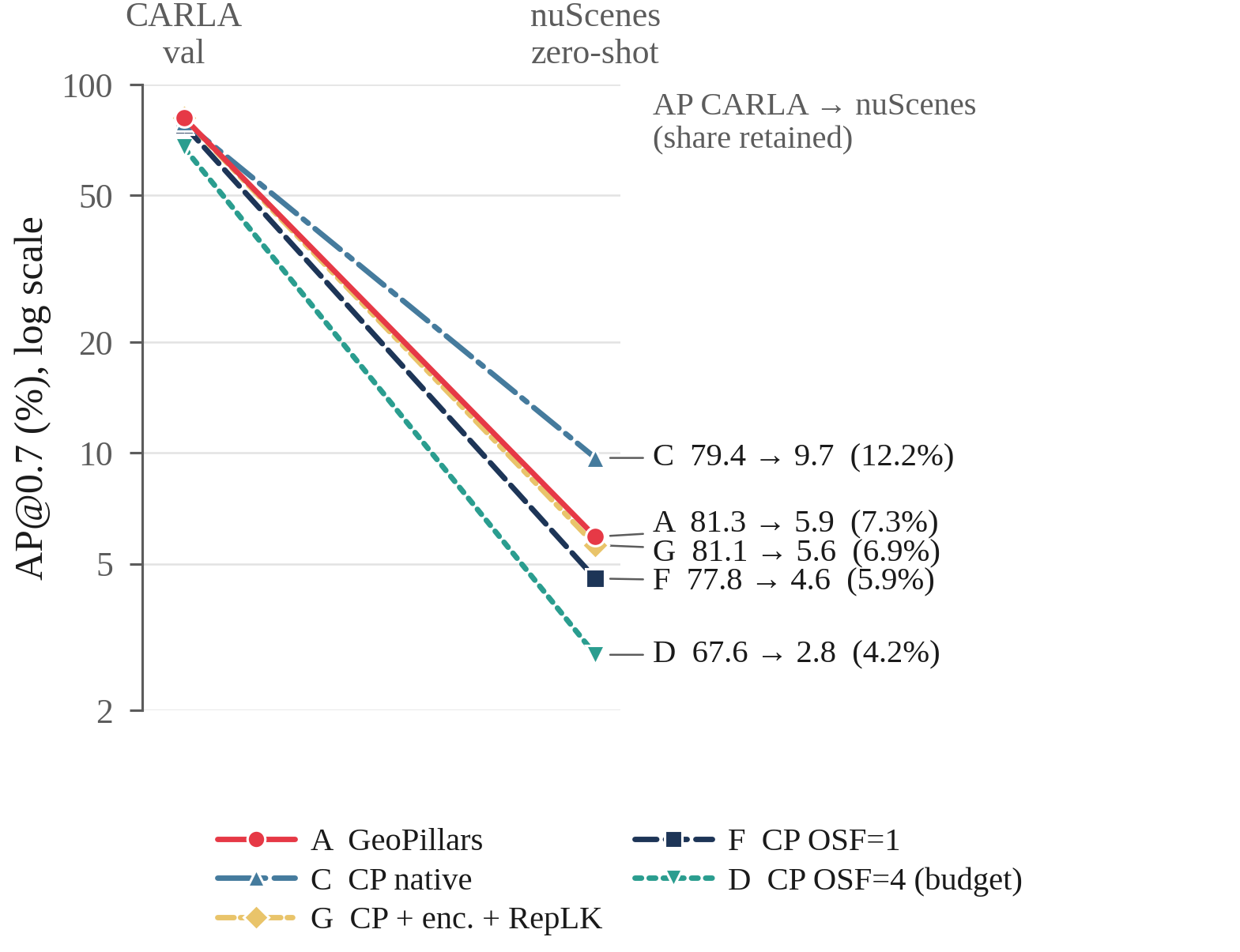}
  \caption{Zero-shot transfer from CARLA to nuScenes, AP@0.7 on a log scale, for five of the seven runs. Retention ranges from 4.19\% (D) to 12.20\% (C) among these and from 3.67\% (E) to 12.20\% over all seven, and the native-budget baseline C retains the most. Among full-resolution runs, A and G, which carry the positional prior, retain 7.28\% and 6.93\% (B, not shown, 8.18\%) against 5.86\% for F, the opposite of what the prior hypothesis predicts.}
  \label{fig:sim2real}
\end{figure}

\subsection{Cost}
\label{subsec:cost}
\begin{figure}[!t]
  \centering
  \includegraphics[width=\columnwidth]{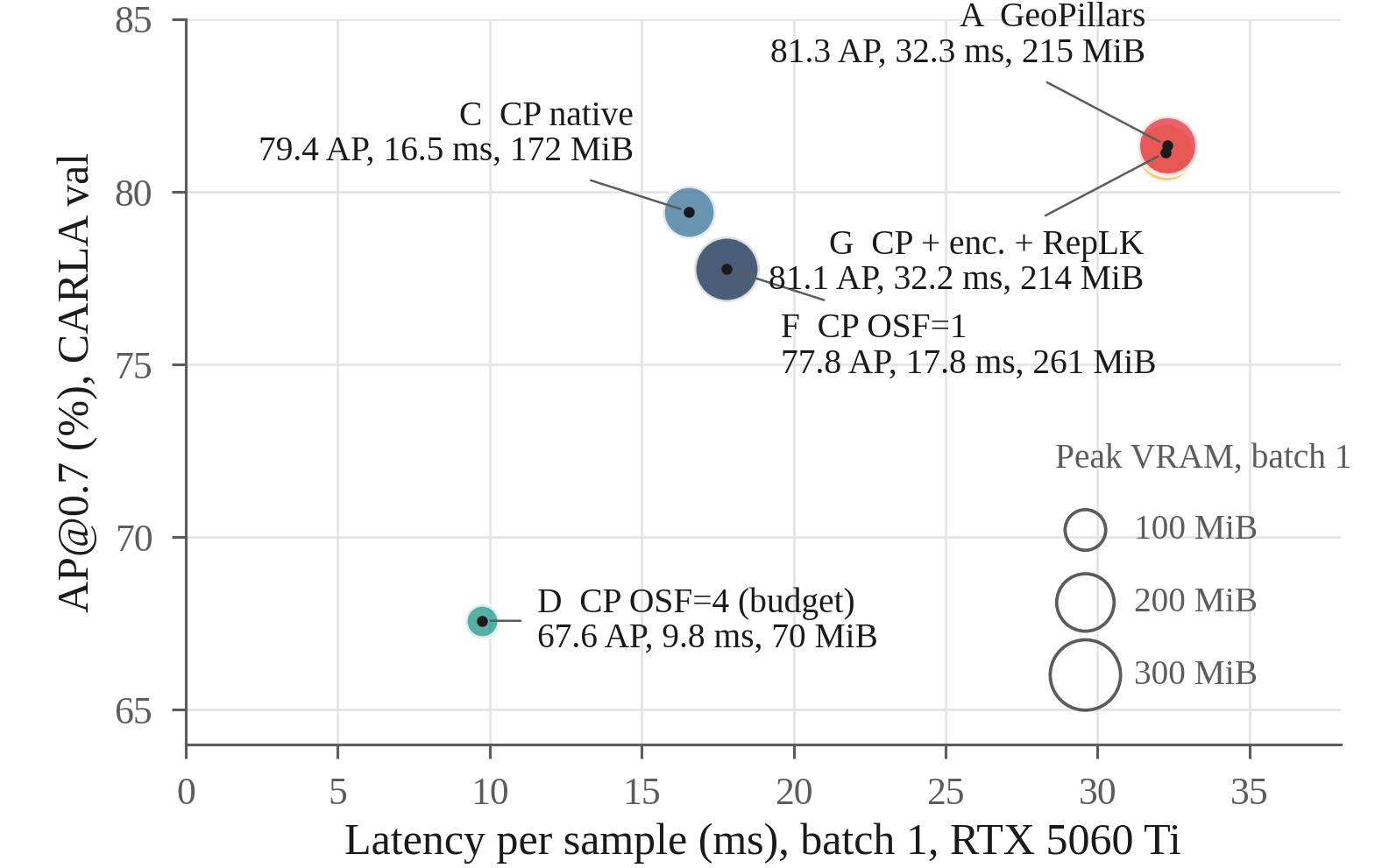}
  \caption{AP@0.7 on the CARLA validation split against latency per sample at batch 1 for five of the seven runs, with marker area proportional to peak memory. B and E are not shown, and G lies almost behind A. GeoPillars (A) is the slowest of the seven runs, at $3.3\times$ the latency of D and $1.95\times$ that of C, although B and G are within 0.2~ms of it.}
  \label{fig:cost}
\end{figure}

Table~\ref{tab:cost} and Fig.~\ref{fig:cost} report inference cost on one GPU at two batch sizes. GeoPillars is the slowest model, at 32.29~ms per sample at batch 1, with B and G, which share its encoder, backbone and upsampler, within 0.2~ms. That is $3.3\times$ the budget-matched baseline and $1.95\times$ the native-budget baseline, for 1.91 AP over the latter. These are unfused figures (Section~\ref{subsec:backbone}).

Batching changes the scale of memory use but not the ranking. F is the heaviest model at both batch sizes, and the ratio between the heaviest and lightest model grows from $3.75\times$ to $6.51\times$. The four full-resolution models need 144 to 192~MiB more than the budget-matched baselines at batch 1 and 3.4 to 4.5~GiB more at batch 24, so batch-1 memory understates the difference that matters in deployment. Batching also moves latency in both directions. The budget-matched baseline D speeds up from 9.76 to 8.63~ms per sample at batch 24, whereas GeoPillars slows from 32.29 to 34.97~ms, which suggests it is already compute-bound at batch 1.

\begin{table}[!t]
  \caption{Inference cost on one RTX 5060 Ti in fp32 with PyTorch's default TF32 convolutions, timed over 500 CARLA frames after 50 warm-up frames at batch 1 and over 12 batches after 3 warm-up batches at batch 24. Peak is \texttt{max\_memory\_allocated} over the timed iterations. A, B and G are unfused (Section~\ref{subsec:backbone}).}
  \label{tab:cost}
  \centering
  \setlength{\tabcolsep}{4pt}
  \begin{tabular}{c r c c c c}
    \hline
        &           & \multicolumn{2}{c}{Batch 1} & \multicolumn{2}{c}{Batch 24}                          \\
    \cline{3-4}\cline{5-6}
    Run & Params    & ms/sample                   & Peak (MiB)                   & ms/sample & Peak (MiB) \\
    \hline
    A   & 7,276,116 & 32.29                       & 215.1                        & 34.97     & 4,319      \\
    B   & 7,276,052 & 32.13                       & 214.7                        & 34.90     & 4,311      \\
    C   & 4,838,987 & 16.54                       & 171.7                        & 16.93     & 3,209      \\
    D   & 4,838,987 & 9.76                        & 69.9                         & 8.63      & 842        \\
    E   & 4,838,923 & 9.53                        & 69.6                         & 8.67      & 836        \\
    F   & 7,050,827 & 17.79                       & 261.1                        & 17.31     & 5,436      \\
    G   & 6,979,284 & 32.23                       & 213.8                        & 34.55     & 4,304      \\
    \hline
  \end{tabular}
\end{table}

\subsection{The Positional Prior at Inference}
\label{subsec:prior_ablation}
Zeroing the positional prior in the trained checkpoint, with nothing else changed, costs 0.82 AP@0.7, from 0.8133 to 0.8051. The model degrades without collapsing. The loss is largest beyond 40~m, at 3.98 points against 2.27 within 20~m and 1.61 between 20 and 40~m, the heading-flip rate rises from 0.80\% to 0.97\%, and the network emits 10.2\% more detections. This fits a learned fill-in for empty space, since the prior supplies all the content of the 94.9\% of cells that hold no points.

The lesion has two limits. It removes the prior from a network trained with it, so it does not show how a network trained without the prior would perform. The matching lesion on nuScenes, which would test whether the prior encodes CARLA scene layout, was not run. What it does establish is an allocation problem, since the prior holds 4.19~M parameters, 87\% of the size of the whole CenterPoint baseline, for 0.82 points.

\subsection{Qualitative}
\label{subsec:qualitative}
\begin{figure*}[!t]
  \centering
  \includegraphics[width=\textwidth]{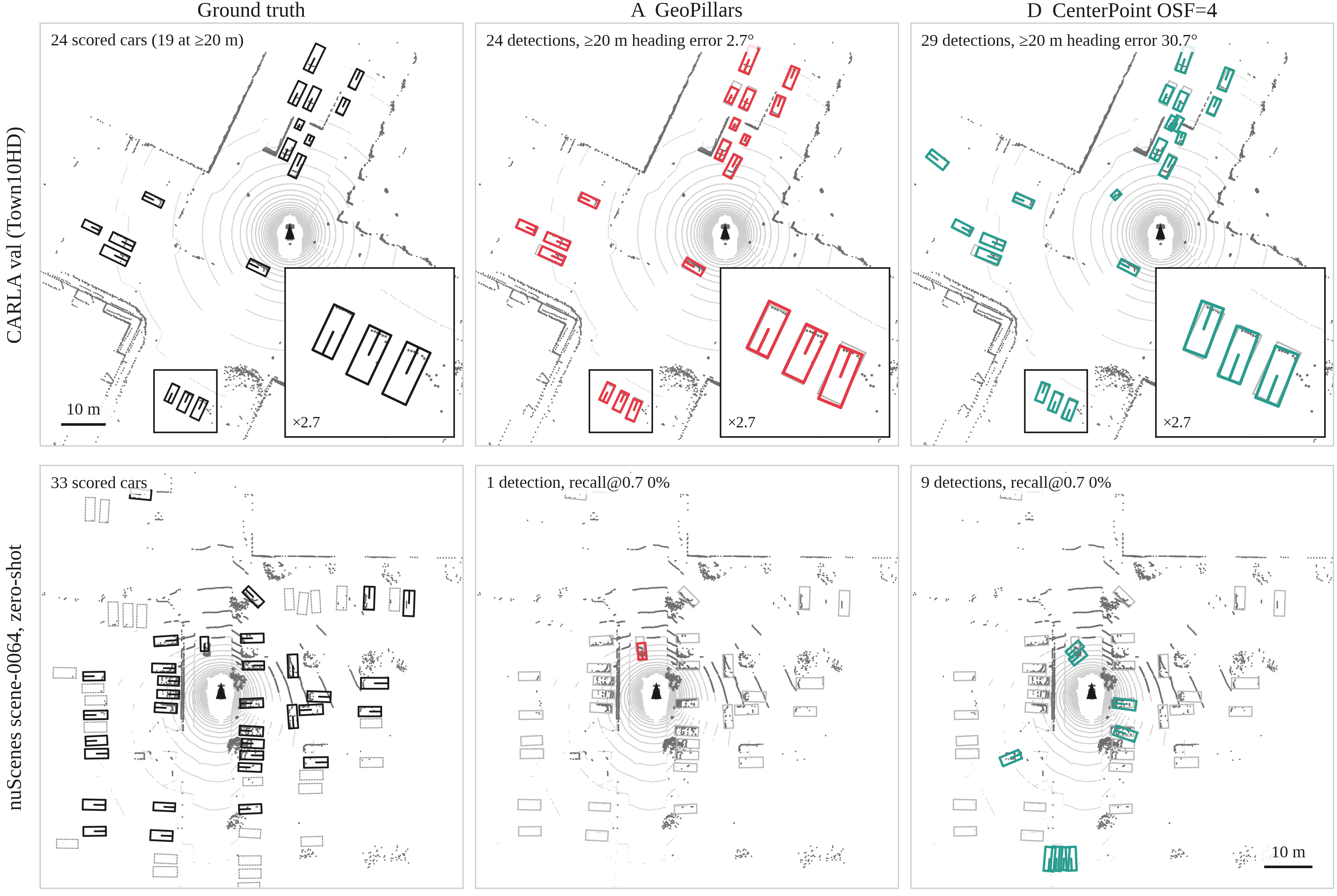}
  \caption{Ground truth, GeoPillars (A) and the budget-matched baseline (D), with ground truth drawn in grey behind the predictions. Top, a CARLA validation frame from Town10HD. Both models match every car at IoU 0.5, but the mean heading error beyond 20~m is $2.7^\circ$ for A and $30.7^\circ$ for D. The inset magnifies three cars near 45~m by $2.7\times$. Bottom, nuScenes scene-0064 zero-shot, where neither model recovers a car at IoU 0.7.}
  \label{fig:qualitative}
\end{figure*}

Fig.~\ref{fig:qualitative} shows one CARLA frame and one nuScenes keyframe for A and D. On the CARLA frame both models match all 24 cars at IoU 0.5 and differ visibly in heading, and at IoU 0.7 A misses 2 cars and D misses 6. Over the whole split, though, A and D differ little in flip rate (0.80\% against 0.92\%) and mainly in false positives, so the frame illustrates a failure mode rather than its frequency. On the nuScenes keyframe both models fire on the scene and neither matches a car at IoU 0.7.

\section{Discussion and Design Lessons}
\label{sec:discussion}
The evaluation did not support the hypothesis behind GeoPillars, and several of its components do not do what they were designed to do. Each of the five lessons below is tied to a measurement in Section~\ref{sec:results}, and we expect them to hold beyond this network. The section closes with an alternative design.

\subsection{Control Head Resolution Before Crediting Architecture}
\label{subsec:lesson_head}
Most of the gain first credited to the architecture came from the head stride. Moving the head from stride 4 to stride 1 recovers 74\% of the gap (Table~\ref{tab:decomposition}). The likely main causes are target assignment and the cell-unit regression target (Table~\ref{tab:radius_configurations}), both of which change with the stride while the model description does not. PillarNeXt~\cite{2023PillarNeXt} had already shown that head resolution matters. What our ladder adds is its size relative to the components it was credited to, about three times the architectural increment. A configuration diff shows only the stride, not the Gaussian radius or the Smooth-L1 transition it implies, so a comparison that changes the stride together with the network mixes the two effects without showing it.

\subsection{Metric Choice Determines the Winner on Degenerate Targets}
\label{subsec:lesson_metric}
The original orientation metric ranked the native-budget baseline worst of seven, and folding heading reversal out of the same quantity ranks it best (Table~\ref{tab:orientation}). The cause is the data. Nearly half of the validation yaws are lane-aligned to within $10^{-6}$~rad (Section~\ref{subsec:dataset}), so regressing the box axis is almost trivial, and an unfolded error mostly measures which of two values the network picked. The orientation measures of nuScenes~\cite{caesar2020nuscenes} and Waymo~\cite{sun2020waymo} both penalise a reversed heading as heavily as they can, nuScenes with an error of $\pi$ and Waymo with zero heading credit. An error that does not fold a symmetry of the target is dominated by that symmetry when the target concentrates on it, so the target distribution should be checked before the metric is chosen.

\subsection{Supervision Drawn From the Network's Own Sweep Is Weakly Privileged}
\label{subsec:lesson_aux}
The normals that supervise the auxiliary branch come from the same sweep the network receives, before the point cap removed part of it. The network still sees the retained points and every neighbouring pillar, so much of what the normals encode may be recoverable from its input, and the LUPI premise~\cite{VAPNIK2009544} was at best weakly met. Two measurements show how little the branch adds. Most of the auxiliary loss falls on background with no object points (Section~\ref{subsec:audit}), so the branch trains mostly on the static scene, and the gate it drives applies the same average gain at every cell (Fig.~\ref{fig:ngfm_mask}). The branch contributes 0.20 points, within run-to-run variation. Deep Sliding Shapes~\cite{song2016deep} reported the analogous result for HHA features, which added nothing to a TSDF volume built from the same depth data.

\subsection{Measure the Constraint and the Solution on the Same Axis}
\label{subsec:lesson_cost}
Of the two input caps, only the point cap binds, as a count of occupied pillars per scan shows (Section~\ref{subsec:sparsity}). We did not apply the same check to the architecture. Input density was restricted to reduce the cost of sparse-LiDAR perception, and the architecture built to cope with the restriction is the slowest model in the study and the second most memory-intensive (Table~\ref{tab:cost}). No single cost indicator is sufficient~\cite{dehghani2022theefficiencymisnomer}, so an input constraint and the architecture built around it should be judged on the same cost axes. On latency and memory, the axes that motivated ours, GeoPillars does worse than the baseline it was meant to beat.

\subsection{Audit Where the Parameters Went}
\label{subsec:pm_prior}
A positional bias added as a coarse spatial signal holds 57.6\% of the model, 87\% of the size of the entire CenterPoint baseline, and is worth 0.82 points (Section~\ref{subsec:prior_ablation}). It does contribute, mostly beyond 40~m, where points are scarce and the prior is the only content of empty cells. It is still a poor allocation, and net of it the F$\to$G rung trades convolutional capacity for a lookup table. The attention block adds little as well. It ends training close to the identity (Section~\ref{subsec:prior}), so to that tolerance the trained encoder is a two-layer perceptron with a max-pool.

We expected the positional bias to explain the transfer failure, because CNNs can exploit absolute position even without an explicit encoding~\cite{kayhan2020translation} and a bias indexed by ego-frame position seemed likely to memorise the layout of the synthetic towns. The evidence points the other way, since the full-resolution runs that carry the bias retain slightly more of their accuracy than the one without it (Table~\ref{tab:zeroshot}). Without the decisive experiment of zeroing the prior on nuScenes, the hypothesis remains untested.

\subsection{What We Would Build Instead}
\label{subsec:lesson_alt}
The most useful result of the cost analysis is one the study did not set out to produce. The native-budget CenterPoint C beats the resolution-matched F by 1.66 AP@0.7, runs faster with less memory at both batch sizes, and has 2.2~M fewer parameters, a lower folded orientation error and more than twice the zero-shot retention. F leads on AP@0.5, false-positive share and flip rate. In this comparison, spending compute on input fidelity worked better than spending it on head resolution.

Against C, GeoPillars buys 1.91 AP points for $1.95\times$ the latency, $1.25\times$ the batch-1 memory and worse transfer. We would keep the large-kernel design only where a lower flip rate is worth that cost, and drop the auxiliary branch. A run without the positional bias is also worth training, since zeroing it raised the flip rate of A from 0.80\% to 0.97\% and the F$\to$G rung does not show where the lower flip rate comes from. Finally, we would split the data by scene and compute any auxiliary target from a source the network cannot reconstruct from its input.

\section{Limitations}
\label{sec:limitations}
Section~\ref{subsec:audit} lists implementation properties that affect specific numbers. The limitations below apply to the study as a whole.

\textbf{Synthetic data and one loosely defined class.} Every model was trained and evaluated in simulation, and none transfers usefully to real data (Section~\ref{subsec:zeroshot}). Only cars were trained and scored, and the car label includes ambulances, police vehicles and static map vehicles (Section~\ref{subsec:dataset}). Detections on the 90,357 non-car boxes count as errors and make up 2.6\% to 5.9\% of false positives in runs A, D and F. Intensity is constant, so the question that motivated runs B and E cannot be answered.

\textbf{Interpolation, not generalisation.} The split shuffles frames rather than scenes, so every CARLA figure is in-distribution, and no scene-disjoint set exists. Velocity is predicted but never supervised, and two definitions in the original evaluator were defective. The original values remain in the released records under their original keys.

\textbf{One run per configuration.} There are no seed repeats. The near-replicates put run-to-run variation at 0.13 to 0.57 points, and the encoder-only and encoder-plus-backbone runs that would split the F$\to$G rung were not trained.

\textbf{Unconverged training on mixed hardware.} All runs stopped at eight epochs with validation curves still rising. Between epochs six and eight F gained 2.19 points against 0.44 for A, so the 3.57-point margin between them was still closing. Four GPU types make training memory and wall-clock time incomparable, although inference cost was measured on one device.

\textbf{Configuration asymmetries.} The Smooth-L1 $\beta$ differs between C and the other six runs and never scales with the head cell (Section~\ref{subsec:head}). The sparsity regime was set by hand, and training and inference keep different points in overflowing pillars, at a cost of 0.03 points.

\textbf{Unmeasured quantities.} We did not measure the folded orientation error or the prior lesion on nuScenes, accuracy by weather or traffic, or the fused model.

\section{Conclusion}
\label{sec:conclusion}
We built GeoPillars to test whether richer pillar geometry, a large receptive field and auxiliary surface-normal supervision could recover accuracy lost to a severe point budget. It reaches 81.33 BEV AP@0.7 on cars and beats CenterPoint at its native, denser budget by 1.91 points, but not for the reasons it was designed. Of its gain over the budget-matched baseline, 74\% comes from the head stride and 24.5\% from a bundled change of encoder, backbone, neck and positional prior, and the auxiliary branch that motivated the design adds 0.20 points. The positional bias holds 57.6\% of the parameters, the attention block ends training near the identity, and no model transfers to real data, where a stock CenterPoint at its native budget fails least.

Whether auxiliary geometric supervision can help sparse detection remains open. This study suggests that a target computed from the network's own sweep cannot settle the question, and we are pursuing it on a standard benchmark.

\appendices

\section{Zero-Shot nuScenes Protocol}
\label{app:zeroshot}
The zero-shot set contains every keyframe of the 170 scenes in nuScenes v1.0-trainval LiDAR blobs 01 and 02, 139 from the official training split and 31 from the validation split, for 6,743 keyframes in all. No model saw nuScenes during training, so split membership does not matter. Each keyframe was converted to a single sweep in the CARLA point format, and its intensity, natively an integer from 0 to 255, was overwritten with 1.0 to match the constant CARLA value.

Ground truth mirrors the CARLA filter. A car counts if it has at least five LiDAR points and its centre lies within 50~m, which keeps 42,705 of the 83,179 car annotations. Detections on the other 40,474 cars are false positives, as detections on unlabelled objects are on CARLA. The evaluator, IoU computation and 40-point interpolation are the same as for CARLA.

The released file \texttt{figures/data/nusc\_gt\_cars.npz} stores all 83,179 car annotations with their sample token, scene name, LiDAR point count and range, together with the 6,743 keyframe tokens, so the set can be reconstructed from the public metadata and the two LiDAR blobs and checked against the reported AP.

\section*{Code and Checkpoints}
The code and configurations are available on \href{https://github.com/NotArsenic/GeoPillars-release}{\textcolor{blue}{GitHub}}. The trained checkpoints are available on \href{https://huggingface.co/GeoPillars/GeoPillars}{\textcolor{blue}{Hugging Face}}.

\section*{Acknowledgments}
This work is the result of the undergraduate research project of C.D., S.S., A.M. and G.P. The dataset constraints, confounds and implementation issues described in Section~\ref{subsec:audit} and Section~\ref{sec:limitations} were found during evaluation, and we report them in full.

\section*{Use of Generative AI}
Generative AI was used as a programming and editing assistant, to write and debug evaluation and plotting scripts, to prepare the code release, and for LaTeX formatting and copy-editing. To verify generated code, the reported CARLA metrics were recomputed from the released checkpoints under one evaluation harness. The released code was checked against the original pipeline on stored CARLA scans. The research idea, the GeoPillars architecture and its design rationale, the experimental design, the training runs, the analysis, the claims, their interpretation and the conclusions are the work of the authors, who reviewed and verified all text and code.

\bibliographystyle{IEEEtran}
\bibliography{references}

\end{document}